\documentclass[10pt,twocolumn,letterpaper]{article}

 \usepackage{cvpr}              
\newcommand{\Tref}[1]{Table~\ref{#1}}
\newcommand{\eref}[1]{Eq.~\eqref{#1}}

\newcommand{\fref}[1]{Fig.~\ref{#1}}
\newcommand{\Fref}[1]{Figure~\ref{#1}}

\usepackage{xcolor}
\newcounter{todos}
\AtEndDocument{\ifnum\value{todos}>0 \PackageWarning{TODOS}{There are \arabic{todos} todos left in this paper! Fix them before submitting the paper!} \fi}

\newcommand{\V}[1]{\ensuremath{\mathbf{#1}}}

\newcommand{\GT}{ground truth\xspace}

\newcommand{\pandoraD}{PANDORA dataset~\cite{dave2022pandora}\xspace}
\newcommand{\rmvp}{RMVP3D\xspace}
\newcommand{\smvp}{SMVP3D\xspace}

\newcommand{\pandora}{PANDORA~\cite{dave2022pandora}\xspace}
\newcommand{\mvas}{MVAS~\cite{cao2023multi}\xspace}
\newcommand{\nero}{NeRO~\cite{liu2023nero}\xspace}
\newcommand{\neus}{NeuS~\cite{wang2021neus}\xspace}
\newcommand{\svolsdf}{S-VolSDF~\cite{wu2023s}\xspace}
\newcommand{\sparseneus}{SparseNeuS~\cite{long2022sparseneus}\xspace}
\newcommand{\nersp}{NeRSP\xspace}

\definecolor{cvprblue}{rgb}{0.21,0.49,0.74}
\usepackage{amsmath,amsfonts,mathtools}
\usepackage[pagebackref,breaklinks,colorlinks,citecolor=cvprblue]{hyperref}
\usepackage{multirow}
\usepackage{pifont}
\usepackage{diagbox}
\usepackage{makecell}
\usepackage{bm}
\usepackage{tocloft}
\usepackage{lipsum}
\usepackage[accsupp]{axessibility}
\newcommand\blfootnote[1]{%
	\begingroup
	\renewcommand\thefootnote{}\footnote{#1}%
	\addtocounter{footnote}{-1}%
	\endgroup
}
\def\paperID{2463} 
\def\confName{CVPR}
\def\confYear{2024}

\title{NeRSP: Neural 3D Reconstruction for Reflective Objects with \\ Sparse Polarized Images}

\author{Yufei Han$^{1\dagger}$~~~Heng Guo$^{1\dagger*}$~~~~Koki Fukai$^{2\dagger}$~~~Hiroaki Santo$^{2}$~~~Boxin Shi$^{3,4}$~~~Fumio Okura$^{2}$\\Zhanyu Ma$^{1}$~~~Yunpeng Jia$^{1}$\\
\small{$^1$Beijing University of Posts and Telecommunications}\\
\small{$^2$Graduate School of Information Science and Technology, Osaka University}\\
\small{$^3$National Key Laboratory for Multimedia Information Processing, School of Computer Science, Peking University}\\
\small{$^4$National Engineering Research Center of Visual Technology, School of Computer Science, Peking University}\\
\small{\texttt{\{hanyufei, guoheng, mazhanyu\}@bupt.edu.cn~~~shiboxin@pku.edu.cn}}\\
\small{\texttt{\{santo.hiroaki, okura, fukai.koki\}@ist.osaka-u.ac.jp~~~xibei156@163.com}}\\
}

\begin{document}
\maketitle
\begin{abstract}
We present NeRSP, a \underline{Ne}ural 3D reconstruction technique for \underline{R}eflective surfaces with \underline{S}parse \underline{P}olarized images.
Reflective surface reconstruction is extremely challenging as specular reflections are view-dependent and thus violate the multiview consistency for multiview stereo. On the other hand, sparse image inputs, as a practical capture setting, commonly cause incomplete or distorted results due to the lack of correspondence matching.
This paper jointly handles the challenges from sparse inputs and reflective surfaces by leveraging polarized images. We derive photometric and geometric cues from the polarimetric image formation model and multiview azimuth consistency, which jointly optimize the surface geometry modeled via implicit neural representation.
Based on the experiments on our synthetic and real datasets, we achieve the state-of-the-art surface reconstruction results with only $6$ views as input. 
\vspace{-1em}
\end{abstract}    
\blfootnote{$^{\dagger}$ Equal contribution. $^{*}$ Corresponding author.}
\blfootnote{Project page: \href{https://yu-fei-han.github.io/NeRSP-project/}{https://yu-fei-han.github.io/NeRSP-project/}}
\section{Introduction}
\label{sec:intro}

Multiview 3D reconstruction is a fundamental problem in computer vision (CV) and has been extensively studied for many years~\cite{hartley2003multiple}.
With the advancement of implicit surface representation~\cite{park2019deepsdf, sitzmann2020implicit} and neural radiance fileds~\cite{mildenhall2020nerf}, recent multiview 3D reconstruction methods~\cite{boss2021neural, zhang2021physg, wang2021neus, yariv2021volume} have made tremendous progress. 
Despite the compelling shape recovery results, most multiview stereo (MVS) methods still rely heavily on finding correspondence between views, which is particularly challenging for reflective surfaces and sparse input views. 

 \newcommand{\imgs}{0.6}
\newcommand{\imgl}{0.4}
\begin{figure}
	\centering
	\Large
	\resizebox{\linewidth}{!}{
	\begin{tabular}{c@{}c@{}c}
		 \includegraphics[width=\imgs\linewidth]{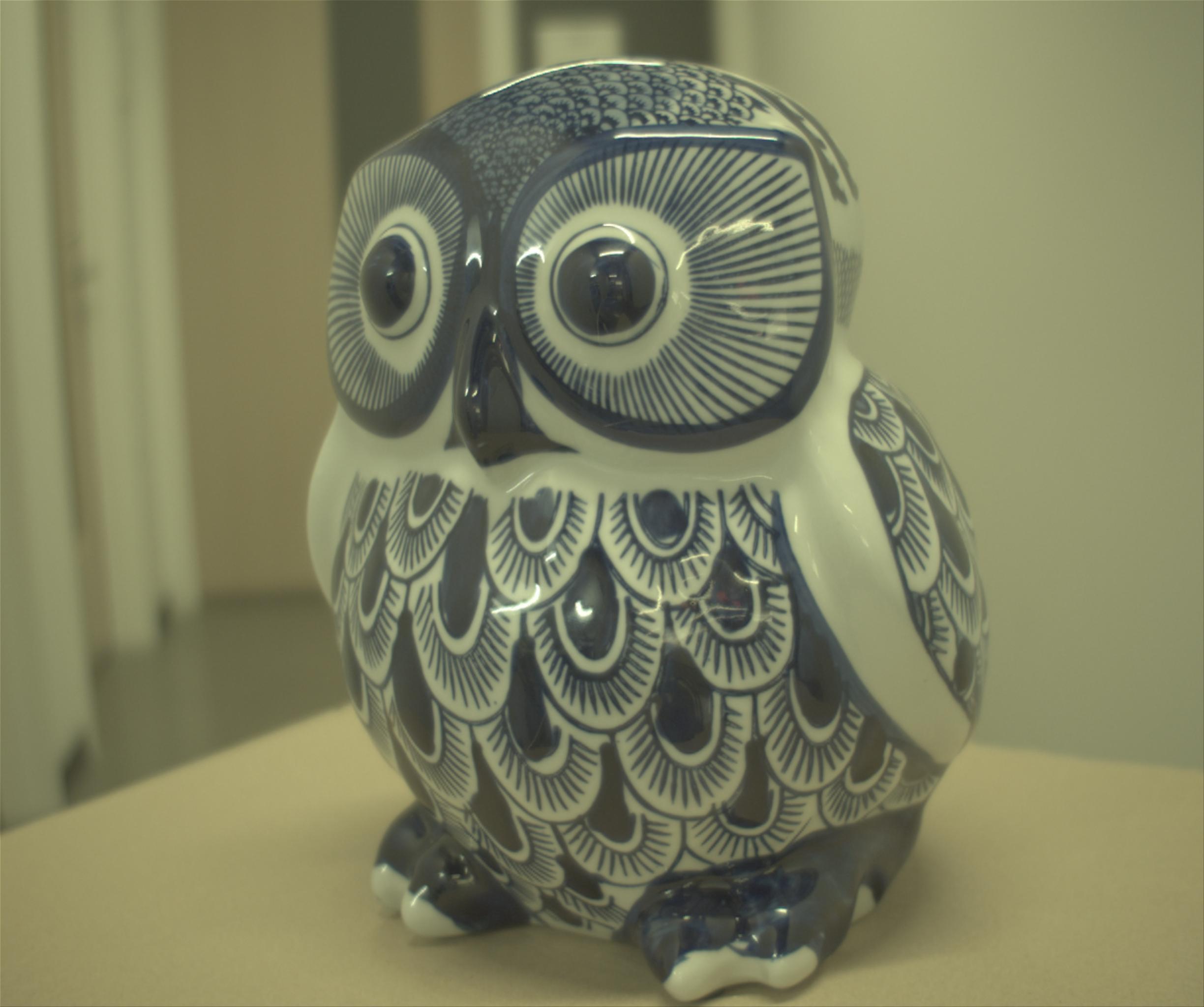}
		& \includegraphics[width=\imgs\linewidth]{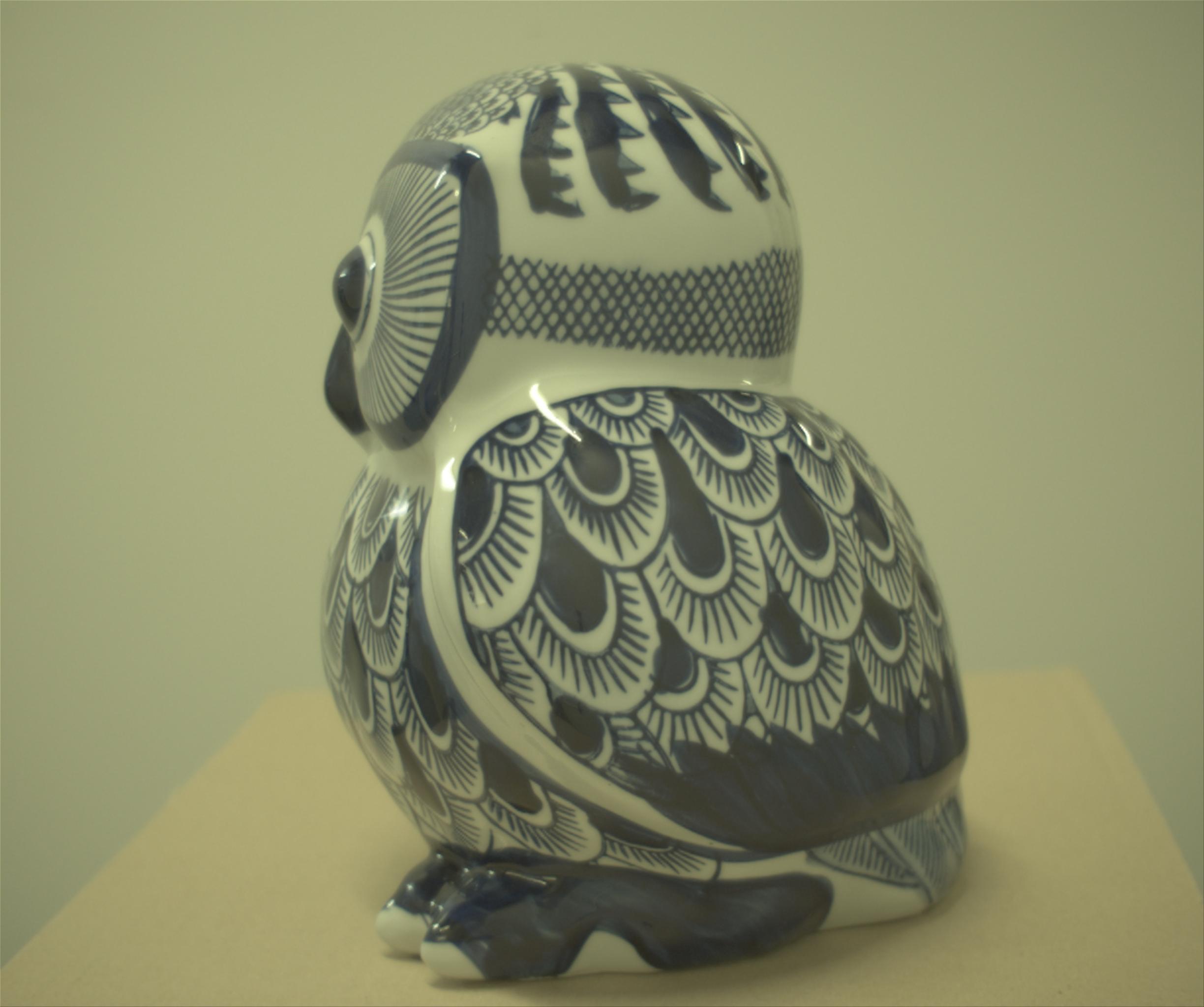}
		& \includegraphics[width=\imgs\linewidth]{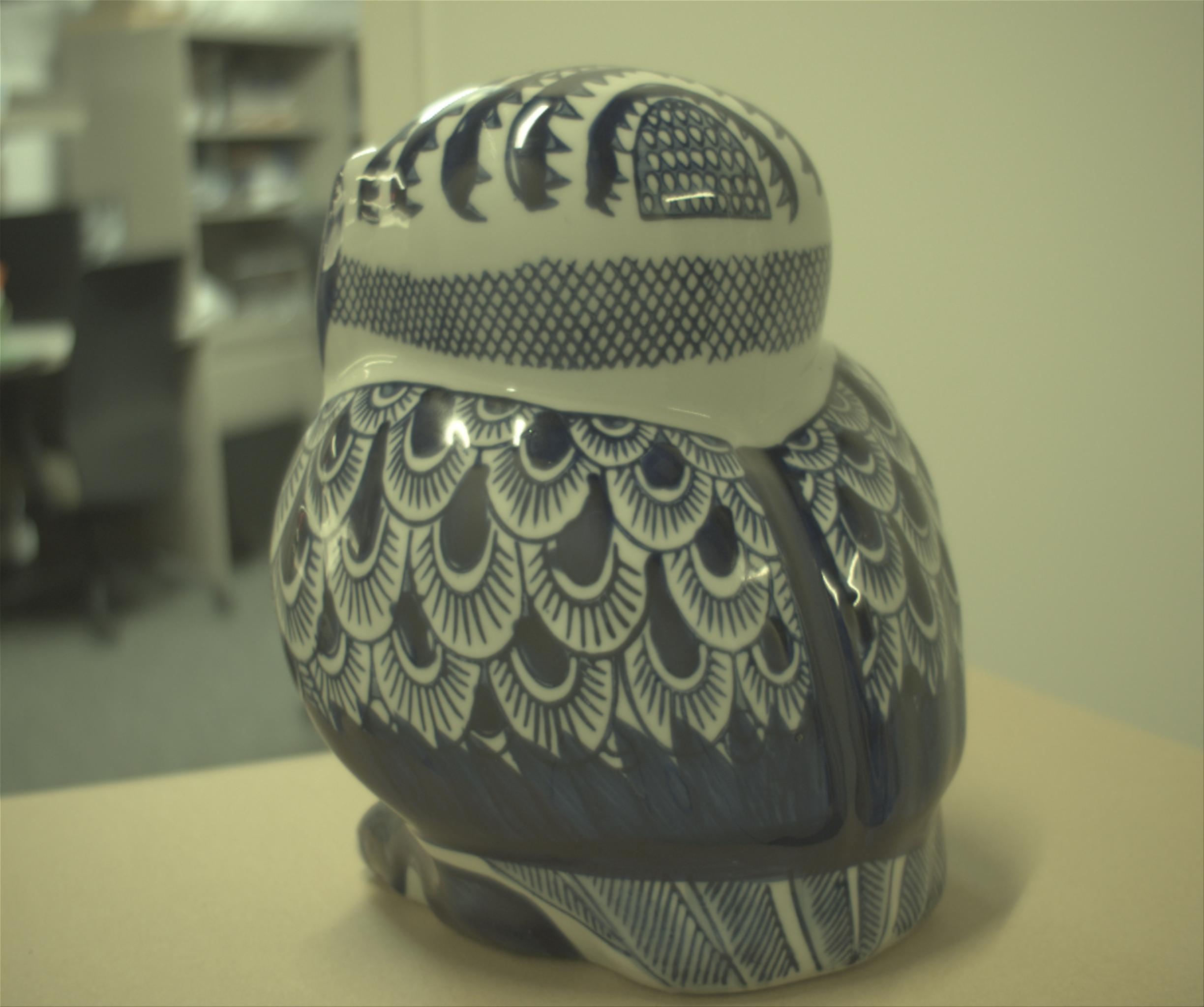}\\
		 \includegraphics[width=\imgs\linewidth]{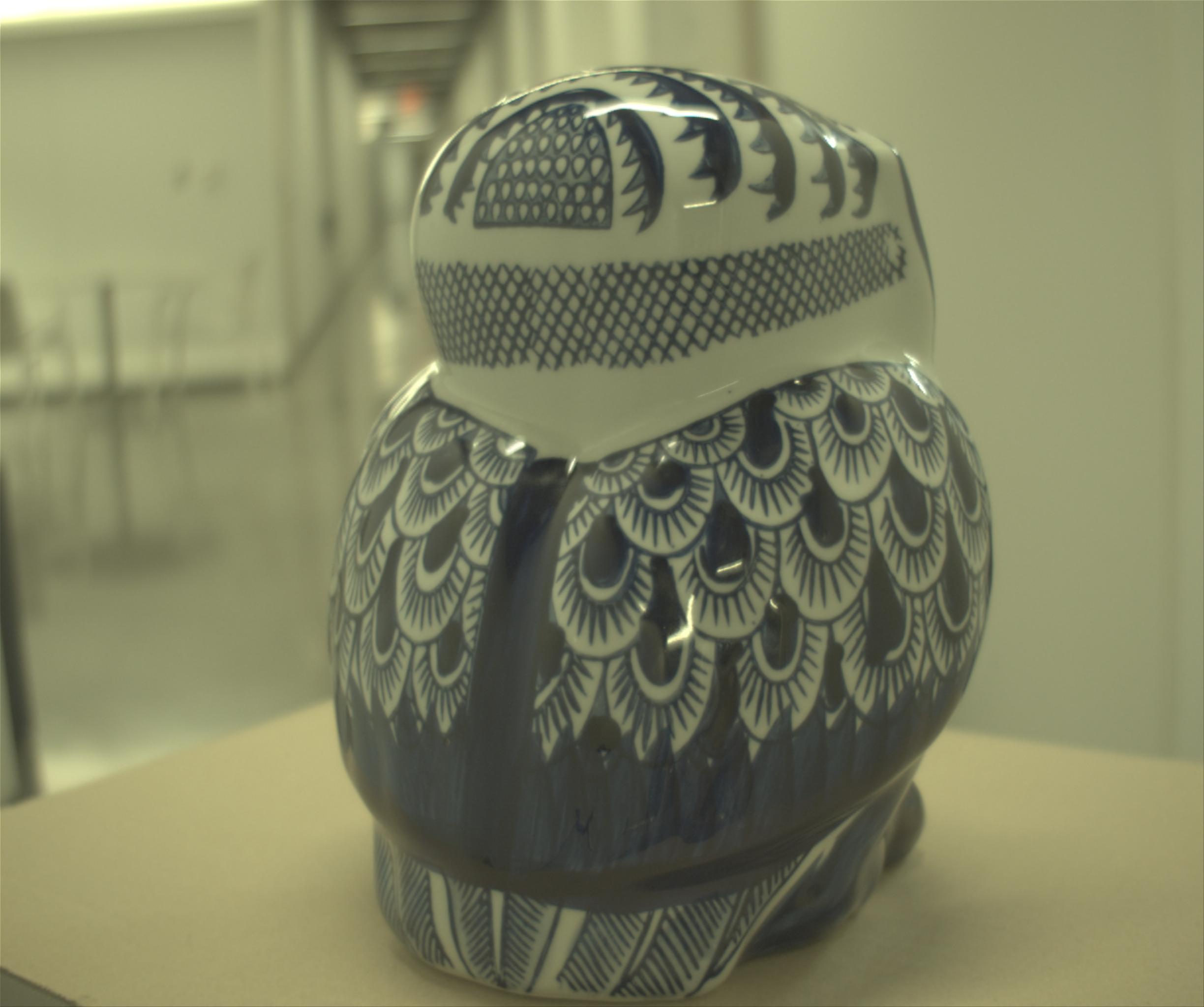}
		& \includegraphics[width=\imgs\linewidth]{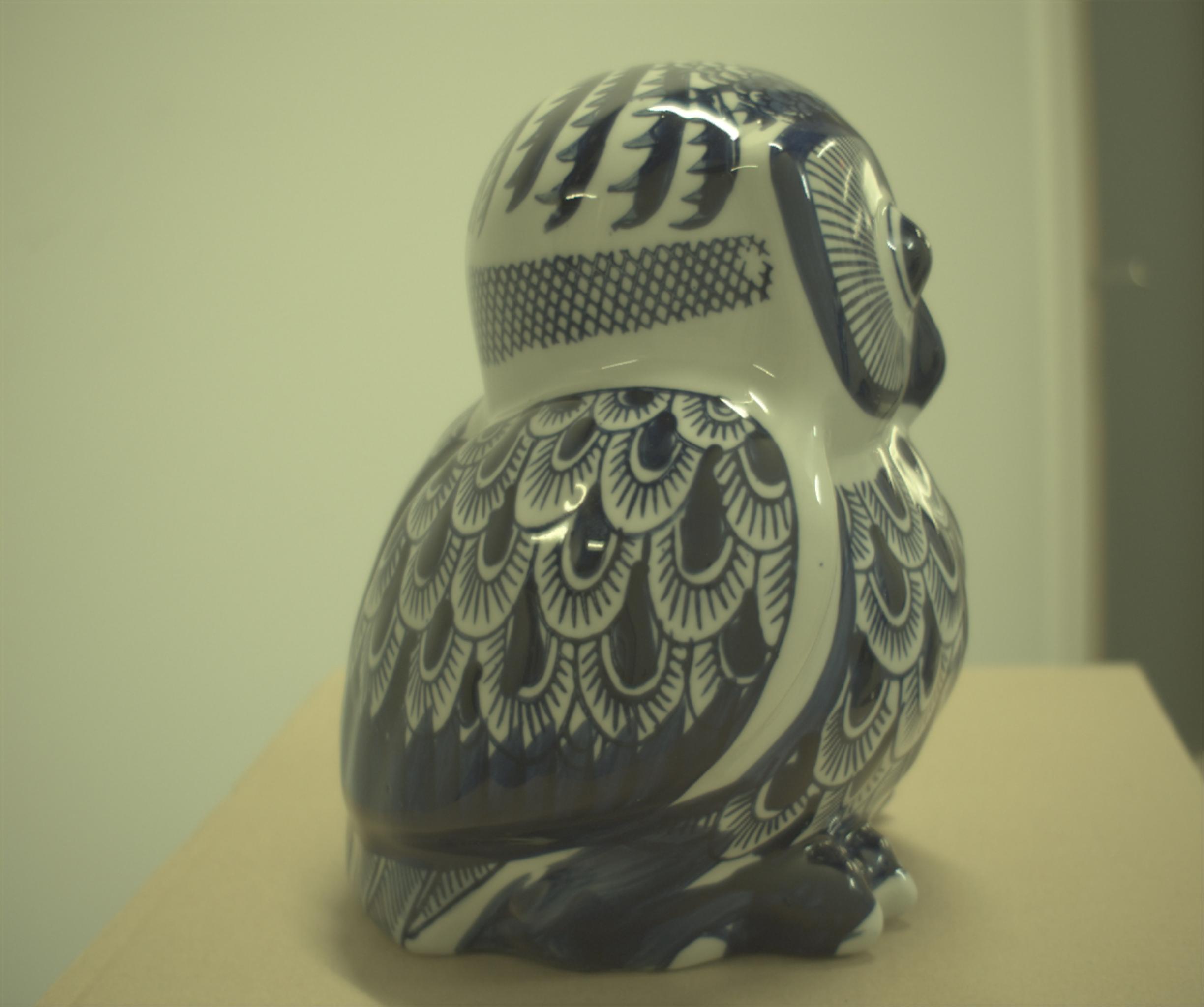}
		& \includegraphics[width=\imgs\linewidth]{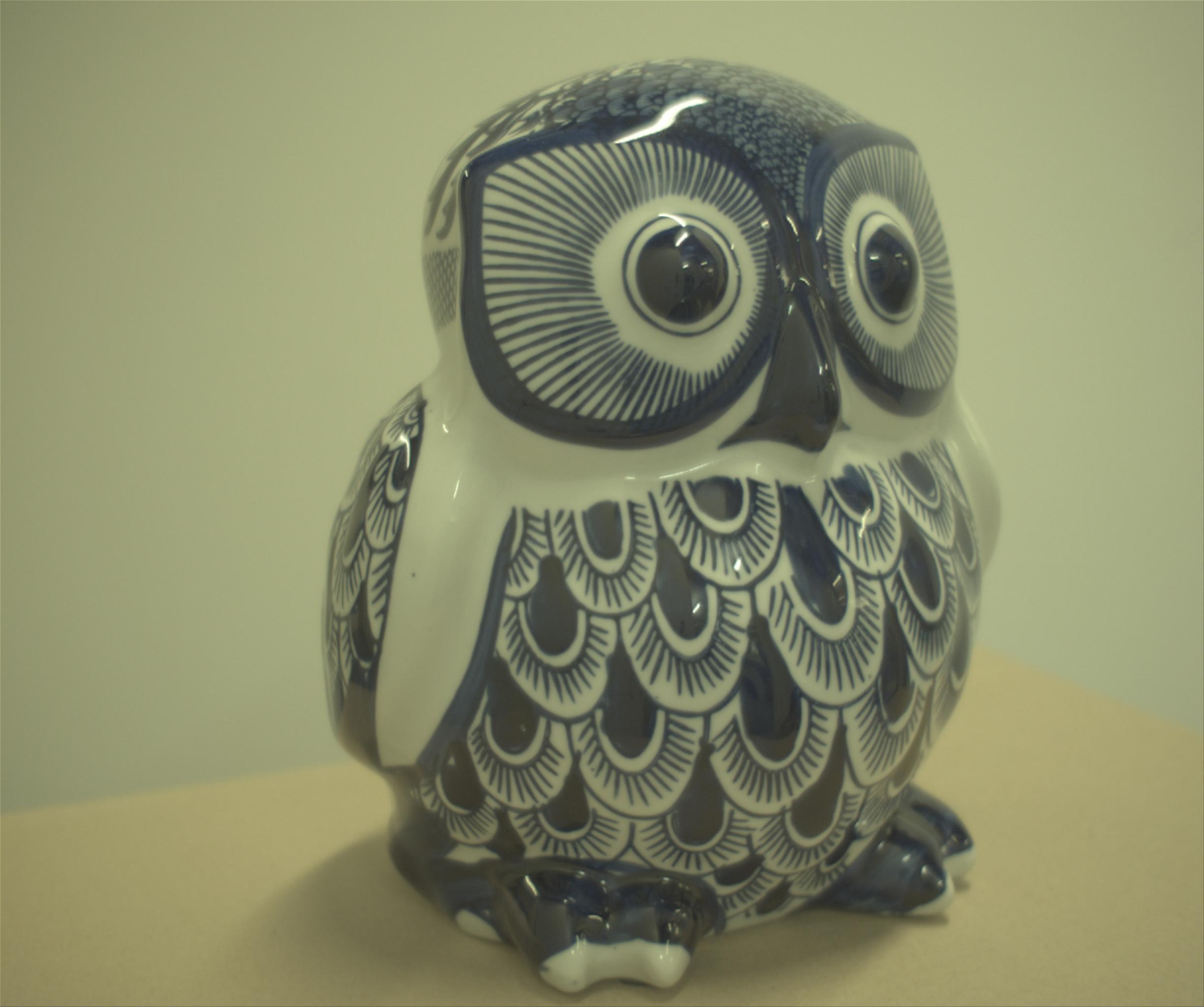} \\
		{\includegraphics[width=\imgl\linewidth]{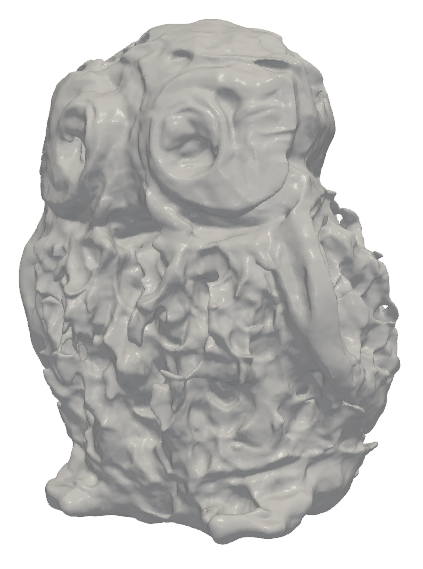}}
		&{\includegraphics[width=\imgl\linewidth]{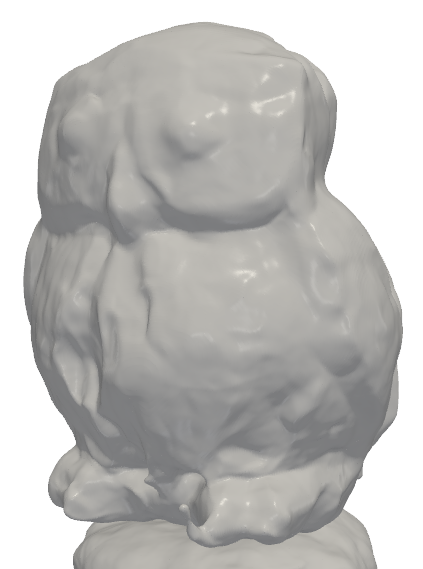}}
		&{\includegraphics[width=\imgl\linewidth]{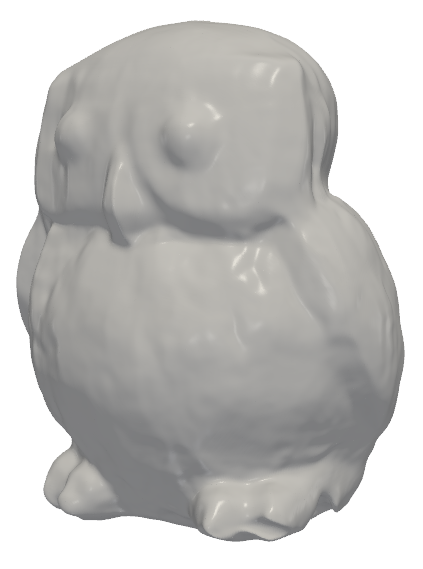}}
		\\
		{\svolsdf}
		&{\pandora}
		& {\nersp~(Ours)}
	\end{tabular}
	}
	\caption{Shape recoveries of a reflective surface from 6 sparse polarized images capturing~(top rows). Our \nersp achieves a better shape reconstruction result compared to existing methods that either address sparse inputs~(\svolsdf) or reflective reflectance~(\pandora).}
	\label{fig:teaser}
	\vspace{-1em}
\end{figure}

For reflective surfaces, the view-dependent surface appearance breaks the photometric consistency assumption used in the correspondence estimation in MVS.
To address this problem, recent neural 3D reconstruction methods~(\eg, Ref-NeuS~\cite{ge2023ref}, NeRO~\cite{liu2023nero}, and \pandora) explicitly model the reflectance and simultaneously estimate the reflectance and environment maps via inverse rendering. However, dense image acquisition under diverse views is required to faithfully handle the additional unknowns besides shape, such as albedo, roughness, and environment map.

From sparse input views, it is often challenging to find sufficient multiview correspondences. 
Especially when representing view-dependent reflectances, it is difficult to disentangle shape from radiance under a limited number of correspondences, leading to shape-radiance ambiguity~\cite{zhang2020nerf}.
Recent neural 3D reconstruction methods for sparse views~(\eg, \svolsdf and \sparseneus) require regularization using photometric consistency, which can be violated for reflective surfaces.

To address both problems, we propose to use sparse polarized images instead of RGB inputs.
Specifically, we propose \nersp, a \underline{Ne}ural 3D reconstruction method to recover the shape of \underline{R}eflective surfaces from \underline{S}parse \underline{P}olarized images.
We use the angle of polarization (AoP) derived from polarized images, which directly reflects the azimuth angle of the surface shape up to $\pi$ and $\pi/2$ ambiguities.
This geometric cue is known to enable multiview shape reconstruction regardless of surface reflectance properties, but the estimated shape based solely on the geometric cue is ambiguous~\cite{cao2023multi} under sparse views settings. 
On the other hand, photometric cue from the polarimetric image formation model~\cite{baek2018simultaneous} helps neural surface reconstruction~(\eg, \pandora) by minimizing the difference between re-rendered and captured polarized images. However, estimated shape based solely on the photometric cue is also ill-posed under sparse inputs due to the shape-radiance ambiguity. 
Unlike the existing polarimetric-based method \pandora considering the photometric cue only, our \nersp shows the integration of both geometric and photometric cues effectively narrows down the solution space for surface shape, shown to be effective in reflective surface reconstruction based on sparse inputs, as visualized in \fref{fig:teaser}.


Besides the proposed \nersp for 3D reconstruction, we also build a \underline{R}eal-world \underline{M}ulti\underline{V}iew \underline{P}olarized image dataset containing $6$ objects with aligned ground-truth (GT) \underline{3D} meshes, named \rmvp. Different from existing datasets such as \pandoraD providing polarized images only, the aligned GT meshes and the surface normals for each view allow a quantitative evaluation of multiview polarized 3D reconstruction.

To summarize, we advance multiview 3D reconstruction by proposing
\begin{itemize}
	\item \nersp, the first method proposing to use the polarimetric information for reflective surface reconstruction under sparse views;  
	\item a comprehensive analysis for the photometric and geometric cue derived from polarized images; and
	\item \rmvp, the first real-world multiview polarized image dataset with GT shapes for quantitative evaluation.
\end{itemize}

\section{Related work}

\paragraph{Multiview 3D reconstruction} has been extensively studied for decades. 
Neural Radiance Fields~(NeRF)~\cite{mildenhall2020nerf, zhang2020nerf, barron2021mip} achieves great success on novel view synthesis in recent years. Inspired by NeRF, neural 3D reconstruction methods~\cite{niemeyer2020differentiable} are proposed, where the surface shape is modeled implicitly via signed distance field~(SDF).
Beginning from DVR~\cite{niemeyer2020differentiable}, the followed-up methods improve the shape reconstruction quality via differentiable sphere tracing~\cite{yariv2020multiview}, volume rendering~\cite{yariv2021volume, oechsle2021unisurf, wang2021neus}, or detail enhanced shape representation~\cite{wang2022hf, li2023neuralangelo}. These methods can achieve convincing shape estimation for diffuse surfaces where photometric consistency is valid across views.
\vspace{-0.8em}
\paragraph{3D reconstruction for reflective surfaces} is challenging as the photometric consistency is invalid. Existing methods~\cite{zhang2021physg, zhang2021nerfactor, boss2021neural} explicitly model the view-dependent reflectance, and disentangle the shape, spatially-varying illuminations, reflectance properties like albedo and roughness. However, the estimates of the above variables are open unsatisfactory as the disentanglement is highly ill-posed. 
\nero proposes using the split-sum approximation of the image formation model and further improves shape reconstruction quality without requiring object masks. However, the above methods typically require dense image capture to guarantee plausible shape recovery results for challenging reflective surfaces.
\vspace{-0.8em}
\paragraph{3D reconstruction with sparse views} is essential for practical scenarios requiring efficient capture. Due to the lack of sufficient correspondence from limited views, the shape-radiance ambiguity cannot be resolved, leading to noisy and distorted shape recoveries. Existing methods address this problem by adding regularizations such as surface geometry smoothness~\cite{niemeyer2022regnerf}, coarse depth prior~\cite{wang2023sparsenerf, deng2022depth}, or frequency control of the positional encoding~\cite{yang2023freenerf}.
Some methods~\cite{chen2021mvsnerf,yu2021pixelnerf,long2022sparseneus} formulate the sparse 3D reconstruction as a conditioned 3D generalization problem where image features pre-trained are used as generalizable priors. 
\svolsdf applies classical multiview stereo method as initialization and regularizes the neural rendering optimization with a probability volume. However, it is still challenging for current methods to recover reflective surfaces accurately.
\vspace{-0.8em}
\paragraph{3D reconstruction using polarized images} has been studied for both single view setting~\cite{miyazaki2003polarization, smith2018height, ba2020deep, baek2018simultaneous, lei2022shape} and multiview setting~\cite{cui2017polarimetric,zhao2022polarimetric, fukao2021polarimetric, ding2021polarimetric, dave2022pandora, cao2023multi}. Unlike RGB images, the AoP from polarized images provides direct cues for surface normal. Single-view shape from polarization~(SfP) techniques benefit from this property and estimate the surface normal under single distant light~\cite{lyu2023shape, smith2018height} or unknown natural light~\cite{ba2020deep, lei2022shape}. Multiview SfP methods~\cite{cui2017polarimetric,zhao2022polarimetric} resolve the $\pi$ and $\pi/2$ ambiguities in the AoP based on the multiview observations. \pandora is the first neural 3D reconstruction method based on polarized images, demonstrated to be effective in recovering surface shape and illumination. \mvas recovers surface shape from multiview azimuth maps, closed related to the AoP maps derived from polarized images. However, these methods do not explore using polarized images for reflective surface reconstruction under sparse shots.

\section{Polarimetric Image Formation Model}
\begin{figure}
	\includegraphics[width=\linewidth]{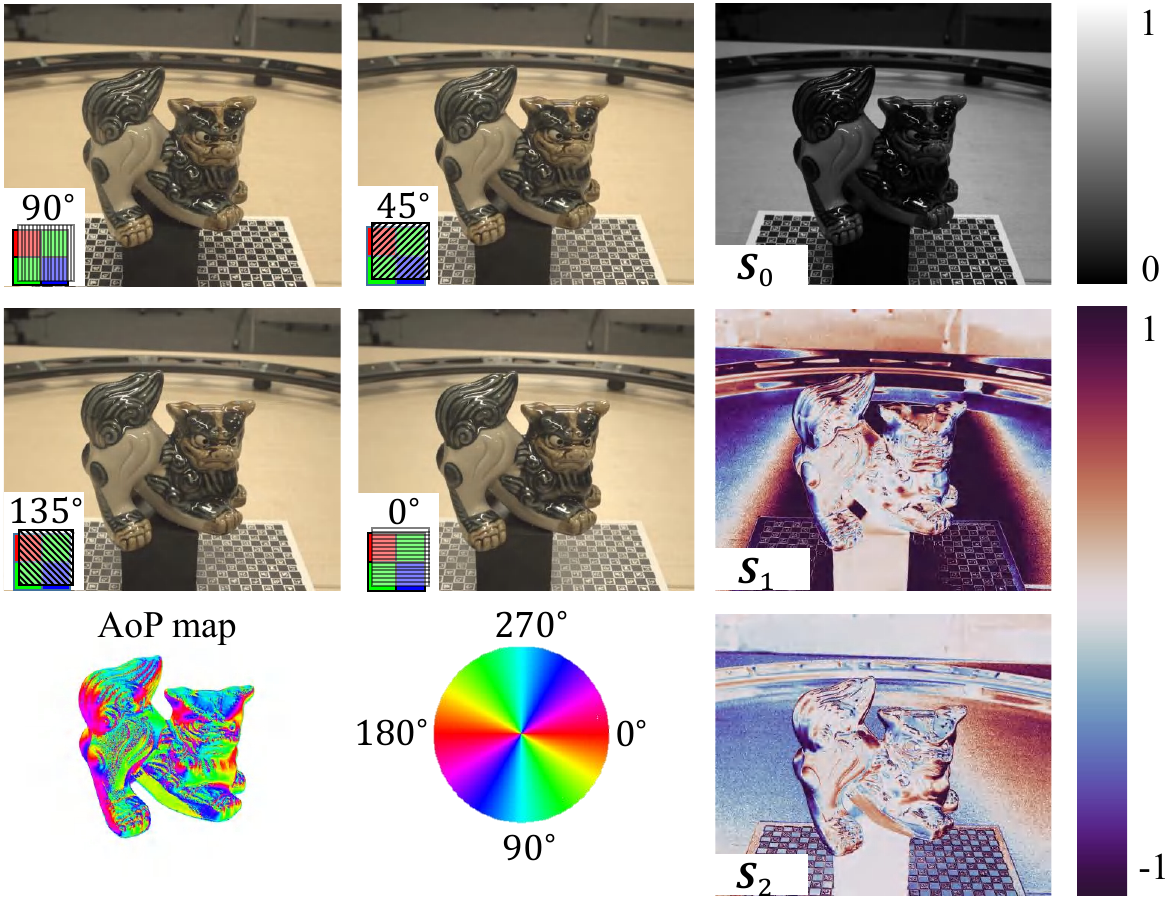}
	\caption{Visualization of polarized images, derived AoP map, and Stokes vectors.}
	\vspace{-0.5em}
	\label{fig:polar_info}
\end{figure}
Before dive into the proposed method, we first introduce polarimetric image formation model and derive the photometric cue and geometric cue in our method.

As shown in \fref{fig:polar_info}, a snapshot polarization camera records image observations at four different polarization angles, with its pixel values denoted as $\{I_{0}, I_{45}, I_{90}, I_{135}\}$. This four images reveal the polarization state of received lights, which is represented as a 4D Stokes vector $\V{s} = [s_0, s_1, s_2, s_3]$ computed as
\begin{eqnarray}
        \label{eq: StokesParameters}
		\begin{dcases}
			s_0 = \frac{1}{2}(I_{0} + I_{45} + I_{90} + I_{135})\\
			s_1 = I_{0} - I_{90}\\
			s_2 = I_{45} - I_{135}.
		\end{dcases}
		\vspace{-0.5em}
\end{eqnarray}
We assume there is no circularly polarized light thus assign $s_3$ to be $0$. The Stokes vector can be used to compute the angle of polarization~(AoP), \ie
\begin{eqnarray}
	\phi_a = \frac{1}{2}\arctan(\frac{s_2}{s_1}).
\end{eqnarray}
Based on the AoP and Stokes vector, we derive the geometric and photometric cue correspondingly.

\subsection{Geometric cue}
Given AoP $\phi_a$, the azimuth angle of the surface can be either $\phi_a + \pi /2$ or $\phi_a + \pi$, known as the $\pi$ and $\pi / 2$ ambiguity depending on whether the surface is specular or diffuse dominant. In this section, we first introduce the geometric cue brought by multiview azimuth map, and then extend it to the case of AoP. 

Following \mvas, for a scene point $\V{x}$, its surface normal $\V{n}$ and the projected azimuth angle $\phi$ in one camera view follow the relationship as
\vspace{-0.3em}
\begin{eqnarray}
	\V{r}_1^\top \V{n} \cos\phi - \V{r}_2^\top \V{n} \sin\phi = 0,
	\label{eq:azi_rotation}
\end{eqnarray}
where $\V{R} = [\V{r}_1, \V{r}_2, \V{r}_3]^\top$ is the rotation matrix of the camera pose. We can further  re-arrange \eref{eq:azi_rotation} to get the orthogonal relationship between surface normal and a projected tangent vector $\V{t}(\phi)$ as defined below,
\vspace{-0.3em}
\begin{eqnarray}
	\V{n}^\top \underbrace{(\cos\phi \V{r}_1 -  \sin\phi \V{r}_2 )}_{\V{t}(\phi)} = 0.
	\label{eq:projected_tan}
\end{eqnarray}
The $\pi$ ambiguity between AoP and azimuth angle can be naturally resolved as \eref{eq:projected_tan} stands if we add $\phi$ by $\pi$. The $\pi/2$ ambiguity can be addressed by using a pseudo projected tangent vector $\hat{\V{t}}(\phi)$ such that
\vspace{-0.3em}
\begin{eqnarray}
	\V{n}^\top \underbrace{(\sin\phi \V{r}_1  +  \cos\phi \V{r}_2 )}_{\hat{\V{t}}(\phi)} = 0.
	\label{eq:projected_tan_psu}
\end{eqnarray}
If one scene point $\V{x}$ is observed by $f$ views, we can stack \eref{eq:projected_tan} and \eref{eq:projected_tan_psu} based on $k$ different rotations and observed AoPs, leading to a linear system 
\vspace{-0.3em}
\begin{eqnarray}
	\V{T}(\V{x}) \V{n}(\V{x}) = \V{0}.
	\label{eq:tangent_consistency}
\end{eqnarray}
We treat this linear system as our geometric cue for multiview polarized 3D reconstruction.

\subsection{Photometric cue}
Assuming the incident environment illumination is unpolarized, the Stokes vector of the incident light direction $\bm{\omega}$ can be represented as 
\vspace{-0.3em}
\begin{eqnarray}
	\V{s}_i(\bm{\omega}) = L(\bm{\omega}) [1, 0, 0, 0]^\top,
	\label{eq:env_stok}
\end{eqnarray}
where $L(\bm{\omega})$ denotes the light intensity. The outgoing light recorded by the polarization camera becomes partially polarized due to the reflection. This process is modeled via a $4 \times 4$ Muller matrix $\V{H}$. Under an environment illumination, the outgoing Stokes vector $\V{s}_o$ can be formulated as the integral of incident Stokes vector multiplicated with the Muller matrix, \ie
\vspace{-0.3em}
\begin{eqnarray}
	\V{s}_o(\V{v}) = \int_{\Omega} \V{H} \V{s}_i(\bm{\omega}) \,\, d\bm{\omega},
	\label{eq:out_stok}
\end{eqnarray}
where $\V{v}$ and $\Omega$ denote the view direction and integral domain.
Following the polarized BRDF~(pBRDF) model~\cite{baek2018simultaneous}, the output Stokes vector can be decomposed into the diffuse and specular parts modeled via $\V{H}_d$ and $\V{H}_s$ correspondingly, \ie
\vspace{-0.3em}
\begin{eqnarray}
	\V{s}_o(\V{v}) = \int_{\Omega} \V{H}_d \V{s}_i(\bm{\omega}) \,\, d\bm{\omega} + \int_{\Omega} \V{H}_s \V{s}_i(\bm{\omega}) \,\, d\bm{\omega}.
	\label{eq:decomp_stok}
\end{eqnarray}
Following the derivation from \pandora, we can furhter formulate the output Stokes vector as
\vspace{0.3em}
\begin{eqnarray}
	\V{s}_o(\V{v}) =   L_d \begin{bmatrix}
		T_o^+ \\
		T_o^- \cos(2 \phi_n) \\
		- T_o^- \sin(2 \phi_n) \\
		0
	\end{bmatrix} +  L_s \begin{bmatrix}
		R^+ \\
		R^- \cos(2 \phi_h) \\
		-R^- \sin(2 \phi_h) \\
		0
	\end{bmatrix},
	\label{eq:photometric_cue}
\end{eqnarray}
where \hbox{$L_d = \int_{\Omega} \rho L(\bm{\omega})  \bm{\omega}^\top \V{n} \, T_i^+T_i^-   \, d\bm{\omega}$} is denoted as diffuse radiance related to surface normal $\V{n}$, Fresnel transmission coefficients~\cite{baek2018simultaneous} $T_{i, o}^+$ and $T_{i, o}^-$, diffuse albedo $\rho$, and the azimuth angle of incident light $\phi_n$. \hbox{$L_s = \int_{\Omega} L(\bm{\omega})  \frac{DG}{4 \V{n}^\top\V{v}}   \,\, d\bm{\omega}$} denotes specular radiance related to Fresnel reflection coefficients~\cite{baek2018simultaneous} $R^+$ and $R^-$, the incident azimuth angle $\phi_h$ \wrt the half vector \hbox{$\V{h} = \frac{\bm{\omega} + \V{v}}{\|\bm{\omega} + \V{v}\|_2^2}$}, and the normal distribution and shadowing term $D$ and $G$ in the Microfacet model~\cite{walter2007microfacet}. Please check supplementary material for more details.

Based on polarimetric image formation model shown in \eref{eq:photometric_cue}, we build the photometric cue.

\section{Proposed method}
Our \nersp takes sparse multiview polarized images, the corresponding silhouette mask of the target object, and camera poses as input, outputs surface shape of the object represented implicitly via SDF. We begin by the discussion on photometric cue and geometric cue in resolving the shape reconstruction ambiguity, followed by the instruction of network structure and loss function of our \nersp.

\subsection{Ambiguity in sparse 3D reconstruction}
\begin{figure}
	\includegraphics[width=\linewidth]{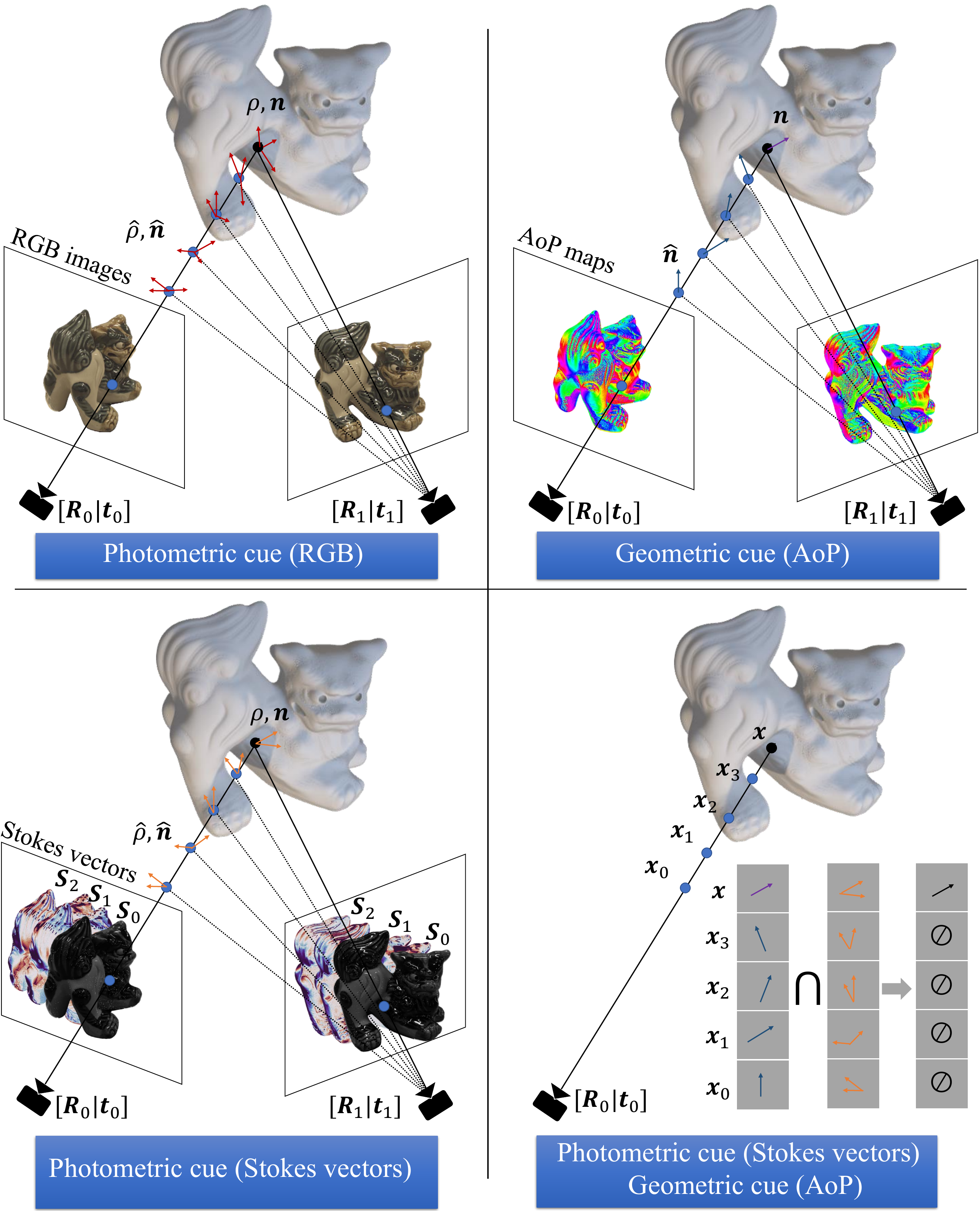}
	\caption{Ambiguity of determining 3D postions under sparse views with geometric and photometric cues.}
	\label{fig:solution_space}
	\vspace{-1.5em}
\end{figure}

The geometric cue and photometric cue play an important role in reducing the solution space of the surface shape under sparse views. As shown in \fref{fig:solution_space}, we illustrate the shape estimation under 2 views with different cues. Given only RGB images as input~(corresponding to the setting in \nero and \svolsdf), different combination of scene point positions, surface normals, and reflectance properties such as albedo can lead to the same image observations, since there are only two RGB measurements for each 3D points along the camera ray. With Stokes vectors extracted from the polarized images, the photometric cue brings $6$ measurements for each 3D points~(Stokes vector has $3$ elements), reducing the surface normal candidates unfit to the polarimetric image formation model. 

On the other hand, based on AoP maps\footnote{AoP is related to the azimuth map discussed in \mvas.} from polarized images, we can uniquely determine the surface normal up to a $\pi$ ambiguity for every scene point along the camera ray. However, it is still ambiguous to find the position where camera ray intersects the surface, unless the third view is provided~\cite{cao2023multi}. Therefore, under sparse views setting~(\eg, 2 views in \fref{fig:solution_space}), determining scene point position based on either geometric or photometric cue remains ambiguous. 

Our method combines these two cues derived from polarized images. As visualized in the bottom-right part of \fref{fig:solution_space}, the correct scene point position should have its surface normal lay in the intersection of normal candidate groups derived from both photometric and geometric cues.
As surface normal at different sampled scene point is uniquely determined by geometric cue, we can easily determine whether the point is on the surface with the aid of photometric cue. 
In this way, we reduce the solution space of sparse-shot reflective surface reconstruction.

\begin{figure*}
	\includegraphics[width=\linewidth]{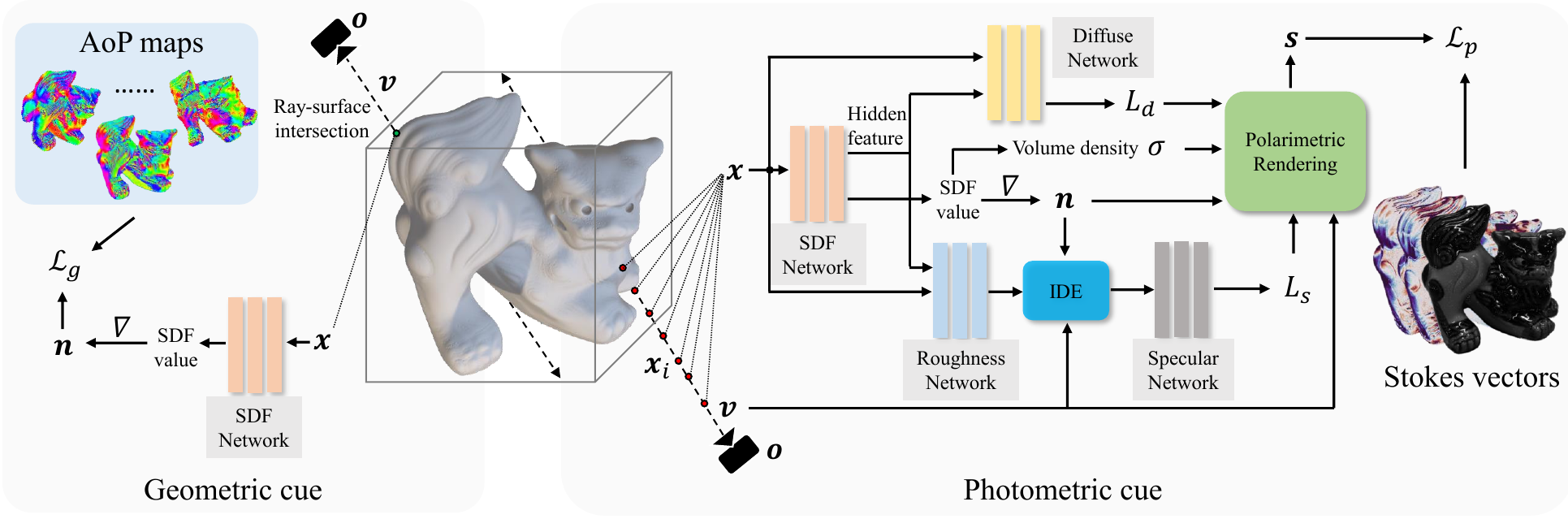}
	\caption{Pipeline of our \nersp.}
	\label{fig:network}
	\vspace{-0.5em}
\end{figure*}

\subsection{\nersp} 
\paragraph{Network structure} As shown in \fref{fig:network}, our \nersp applies a similar network structure with \pandora originally derived from Ref-NeRF~\cite{verbin2022ref}. For a light ray emitted from camera center $\V{o}$ with the direction $\V{v}$, we sample a point on the ray with travel distance $t_i$, its location is denoted at $\V{x}_i = \V{o} + t_i\V{v}$. Following the volume rendering used in NeRF~\cite{niemeyer2022regnerf}, the observed Stokes vector $\V{s}(\V{v})$ can be integrated by the volume opacity $\sigma_i$ and the Stokes vectors at the sampled points along the ray, \ie
\begin{eqnarray}
	\V{s}(\V{v}) = \sum_{i=1}^{n} W_i \V{s}_o(\V{x}_i, \V{v}) \sigma_i,
\end{eqnarray}
where $W_i = \prod_{j=1}^{i-1}(1 - \sigma_j)$ denote the accumulated transmittance of a sampled point.

Motivated by recent neural 3D reconstruction method \neus, we derives the volume opacity from a SDF network and also extract the surface normal from the gradient of the SDF. To compute $\V{s}_o(\V{x}_i, \V{v})$ at sampled points, we follows the polarmetric image formation model in \eref{eq:photometric_cue}. Specifically, the diffuse radiance $L_d$ is related to the diffuse albedo and Fresnel transmission coefficients, which depends on the scene positions but invariant  to the view direction. Therefore, we use a diffuse radiance network to map $L_d$ from the feature of each scene point. The specular radiance $L_s$ is related to the specular lobe determined by the view direction, surface normal, and the surface roughness. We therefore use a RoughnessNet to predict surface roughness. Together with the camera view direction and predicted surface normal, we estimate the specular radiance $L_s$ following the integrated positional encoding module proposed by Ref-NeRF~\cite{verbin2022ref}. Combining $L_d$ and $L_s$, we reconstruct the observed Stokes vector following \eref{eq:photometric_cue}.

\vspace{-0.5em}
\paragraph{Loss function} The photometric loss is defined as the L1 distance between the observed $\hat{\V{s}}(\V{v})$ and reconstructed Stokes vectors $\V{s}(\V{v})$, \ie,
\begin{eqnarray}
	\mathcal{L}_p = \sum_{\V{v} \in \mathcal{V}}  \| \V{s}(\V{v}) - \hat{\V{s}}(\V{v})   \|_1,
\end{eqnarray}
where $\mathcal{V}$ denotes all the camera rays casted within object masks at different views.

For the geometric loss. we first find the 3D scene point $\V{x}$ along the camera ray $\V{v}$ until touching the surface and then locate the projected 2D pixel positions at different views. The geometric loss is defined based on the \eref{eq:tangent_consistency}, \ie, 
\begin{eqnarray}
	\mathcal{L}_g = \sum_{\V{x} \in \mathcal{X}}  \| \V{T}(\V{x})\V{n}(\V{x})   \|_2^2,
\end{eqnarray}
where $ \mathcal{X}$ denotes all the ray-surface intersections inside the object masks at different views.

Besides the photometric and geometric loss, we add mask loss supervised by the object masks and the Eikonal regularization loss. The mask loss is defined as 
\begin{eqnarray}
	\mathcal{L}_m =\sum {\rm BCE}({M}_k, {O}_k),
\end{eqnarray}
where $O_k = \sum_{i=1}^{n} W_{k, i} \sigma_{k, i}$ represents the predicted mask at $k$-th camera ray, whose GT mask value is denoted as $M_k$. BCE represents binary cross entropy loss. 

The Eikonal loss is defined as 
\begin{eqnarray}
	\mathcal{L}_e = \frac{1}{nf}\sum_{i,k}(\|\V{n}_{i, k} \|  - 1)^2,
\end{eqnarray}
where $\V{n}_{i, k}$ is the surface normal derived from the SDF network at $i$-th sampled point along $k$-th camera ray.

Our \nersp is supervised by the combination of the above loss terms, \ie
\begin{eqnarray}
	\mathcal{L} = \mathcal{L}_p + \lambda_g \mathcal{L}_g + \lambda_m \mathcal{L}_m  + \lambda_e \mathcal{L}_e,  
\end{eqnarray} 
where $\lambda_e, \lambda_m$ and $\lambda_p$ are the coefficients for the corresponding loss terms.

\begin{figure*}
	\includegraphics[width=\linewidth,clip={0px,20px,0px,0px}]{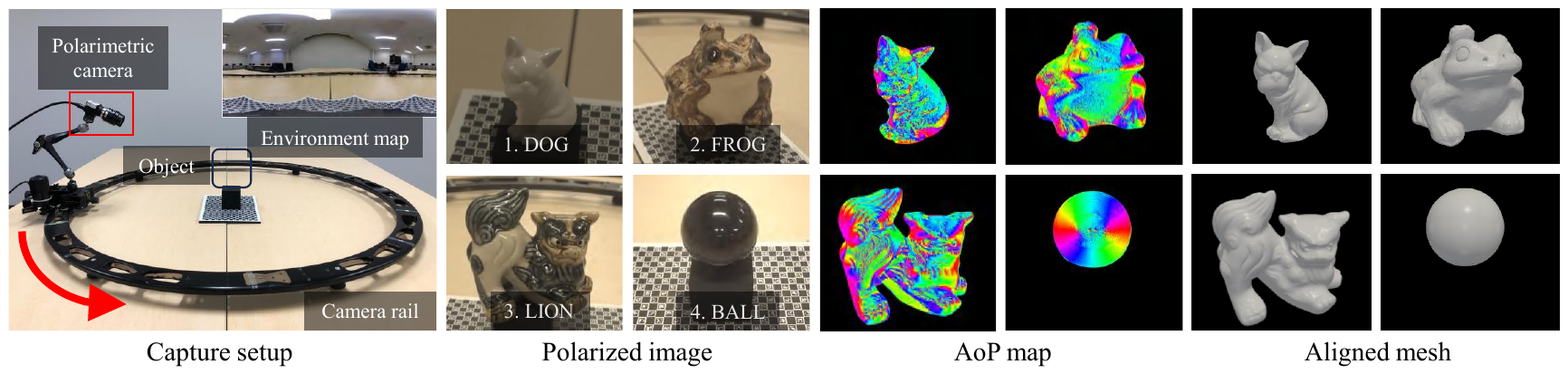}

	\caption{Capture setup and overview of our real-world multiview polarized image dataset \rmvp.}
	\label{fig:rmvp}
	\vspace{-1em}
\end{figure*}

\subsection{\rmvp Dataset}
To quantitatively evaluate the proposed method, we capture a \underline{R}eal-world \underline{M}ultiview \underline{P}olarized image dataset with aligned ground truth meshes. 
\Fref{fig:rmvp}~(left) illustrates our capturing setup, which includes a polarimetric camera, FLIR BFS-U3-51S5PC-C, equipped with a $12$ mm lens and a rotation rail. We use OpenCV for demosaicing the raw data and obtain $1224 \times 1024$ color images with polarizer angles at $0$, $45$, $90$, and $135$ degrees. During the data capture, we place target objects at the center of the rail, and capture 60 images per object by manually moving the camera. We collect $4$ objects as targets: {\sc Dog}, {\sc Frog}, {\sc Lion}, and {\sc Ball}, as shown in \fref{fig:rmvp}~(middle).
For the quantitative evaluation, we adopt a laser scanner Creaform HandySCAN BLACK with the accuracy of $0.01$~mm to obtain the ground truth mesh.
To align the mesh to the captured image views, we first apply \pandora to estimated a reference shape using all available views and then align the scanned mesh to the estimated one via ICP algorithm~\cite{besl1992method}. Besides the ground-truth shapes and multiview images, we also capture the environment map using a 360-degree camera THETA Z1, benefiting quantitative evaluations on the illumination estimation for related neural inverse rendering works.

\section{Experiments}

We evaluate \nersp with three experiments: 1) comparison with existing multiview 3D reconstruction methods quantitatively on synthetic dataset; 2) ablation study on the contribution of geometric and photometric loss terms 3) qualitative and quantitative evaluations on real-world datasets. We also provide the BRDF and novel view results in the supplementary material.

\newcommand{\imgw}{0.2}
\begin{figure}
	\begin{tabular}{@{}c@{}@{}c@{}@{}c@{}@{}c@{}@{}c@{}}
		{\sc Hedgehog } & {\sc Squirrel} & {\sc Snail} & {\sc David} & {\sc Dragon}
		\\
		\includegraphics[height=\imgw\linewidth]{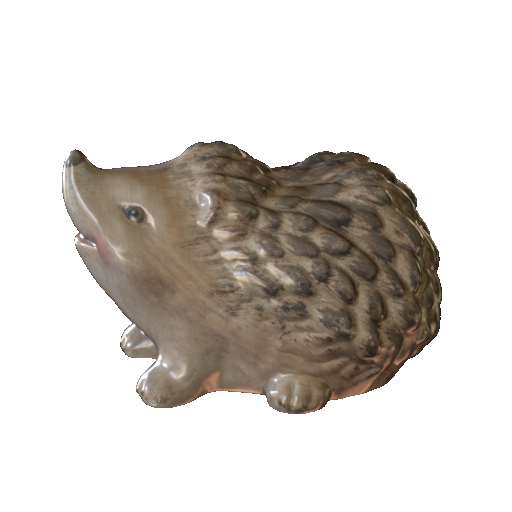}
		&\includegraphics[height=\imgw\linewidth]{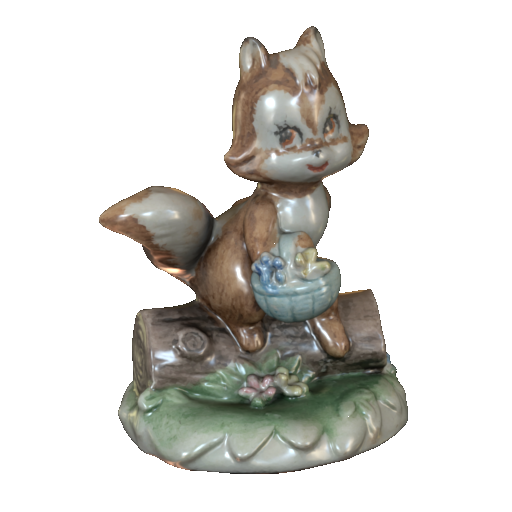}
		&\includegraphics[height=\imgw\linewidth]{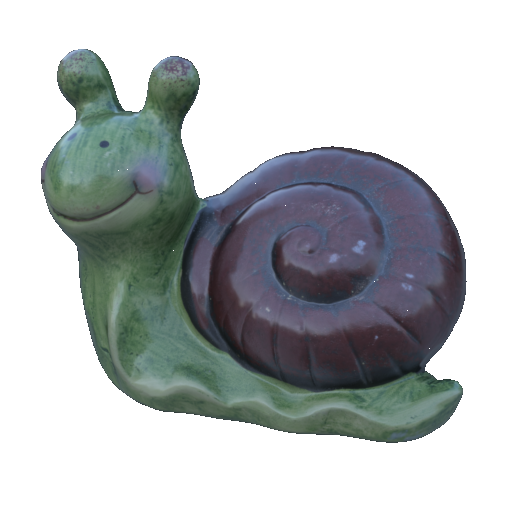}
		&\includegraphics[height=\imgw\linewidth]{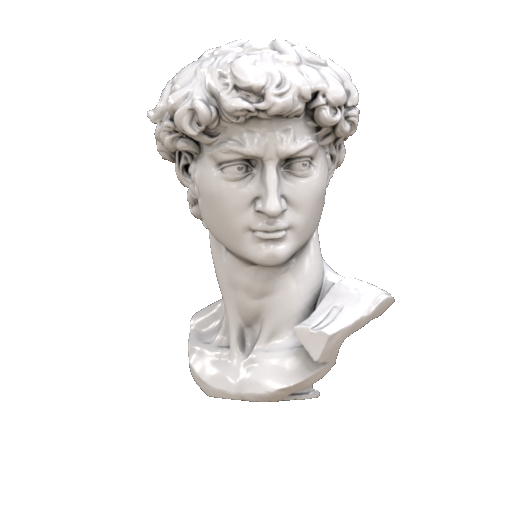}
		&\includegraphics[height=\imgw\linewidth]{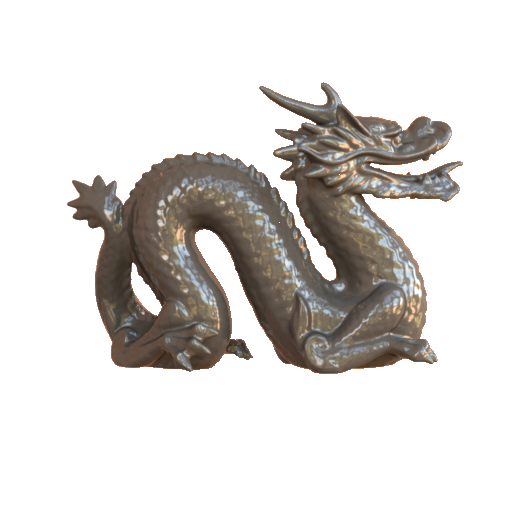}
		\\
		\includegraphics[height=\imgw\linewidth]{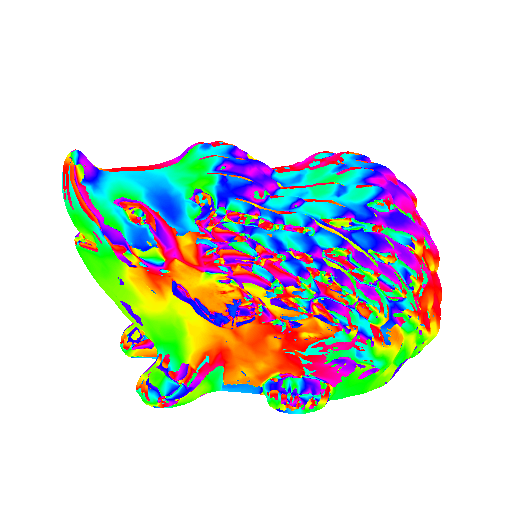}
		&\includegraphics[height=\imgw\linewidth]{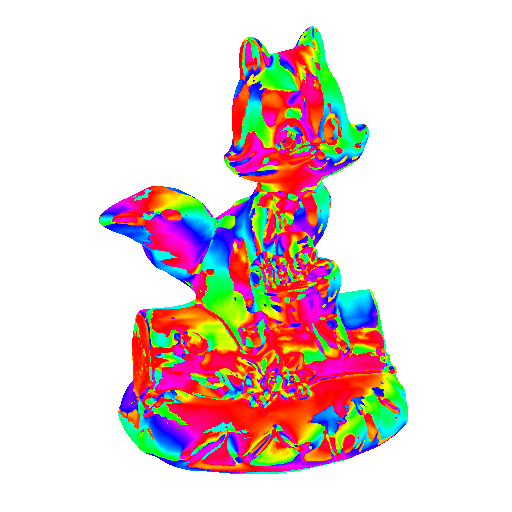}
		&\includegraphics[height=\imgw\linewidth]{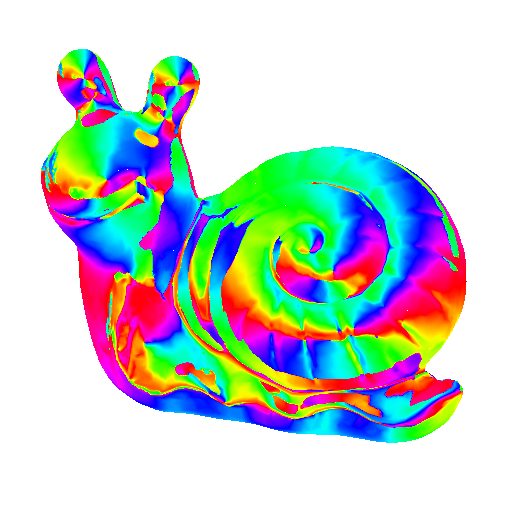}
		&\includegraphics[height=\imgw\linewidth]{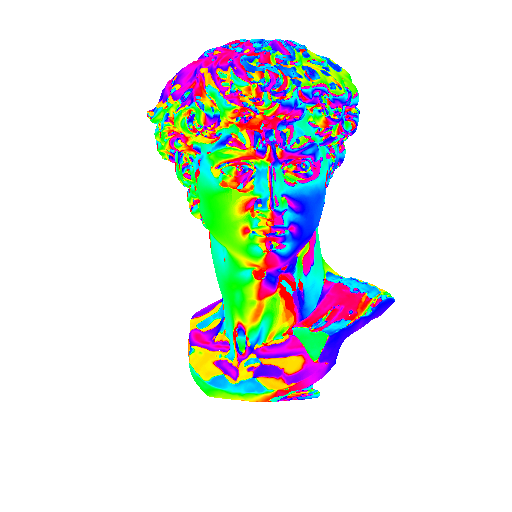}
		&\includegraphics[height=\imgw\linewidth]{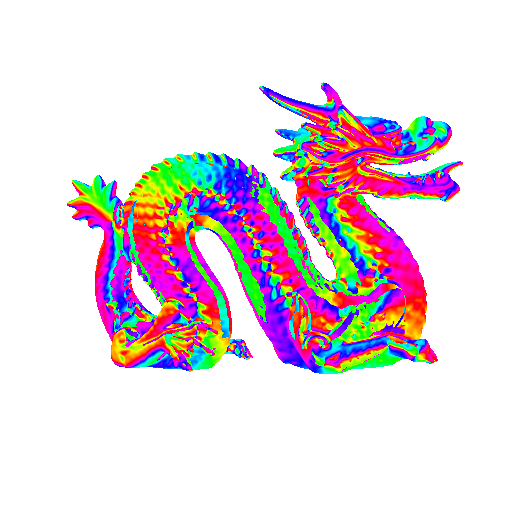}
		\vspace{-1.3em}
	\end{tabular}
	\caption{Overview of synthetic dataset \smvp. Top and bottom rows show image observations and the corresponding AoP maps.}
	\label{fig:synthetic_data}
	\vspace{-1em}
\end{figure}

\subsection{Datasets \& Baselines}
\paragraph{Dataset.} We prepare two real-world datasets: the \pandoraD and our proposed \rmvp, where \pandoraD is only used for qualitative evaluation as the \GT meshes are not provided. 
We also prepare a synthetic multiview polarized image dataset \smvp with Mitsuba rendering engine~\cite{jakob2010mitsuba}, which contains $5$ objects with spatially-varying and reflective reflectance, as visualized in \fref{fig:synthetic_data}. The objects are illuminated by environment maps\footnote{https://polyhaven.com. Retrieved March, 2024.} and captured by $6$ views randomly distributed around the objects.
Besides rendered polarized images, we also export the stokes vectors, GT surface normal maps, and AoP maps for each object.
\vspace{-0.8em}
\paragraph{Baselines.} Our work solves multiview 3D reconstruction for reflective surfaces based on sparse polarized images. Therefore, we choose the state-of-the-art 3D reconstruction methods targeting reflective surfaces \nero and sparse views \svolsdf. The above two methods are based on RGB image inputs. For multiview stereo based on polarized images, we select \pandora and \mvas as our baselines. \nero does not require silhouette masks as input. For a fair comparison, we remove the background in the RGB images with the corresponding masks before inputting to \nero. To compare different methods, we apply Chamfer distance~(CD) between the estimated and the GT meshes, and the mean angular error~(MAE) between the estimated and the GT surface normals at different views as our evaluation metrics. 

\subsection{Shape recovery on synthetic dataset}
As shown in \Tref{table:shape_est_syn}, we summarize the shape estimation error of existing methods and ours on \smvp. Our method achieves the smallest Chamfer distance along all of the $5$ synthetic objects. Based on the visualized shape estimates shown in \fref{fig:syn_shape_vis}, \nero and \svolsdf cannot accurately recover surface details as highlighted in the closed-up views. One possible reason is that the disentanglement of the shape and reflective reflectance from the sparse images is too challenging for these methods based on only RGB information. \mvas and \pandora address the geometric and photometric cues of the polarized images, separately. However, the reconstructed reflective surface shapes are still unsatisfactory due to the ambiguities in geometric and photometric cues under the sparse views setting. As highlighted in the closed-up views, benefiting from both geometric and photometric cues, our method reduces the solution space of shape estimation, leading to the most reasonable shape recoveries compared with the GT shapes.

\begin{table}
	\caption{Comparison on shape recoveries on synthetic dataset evaluated by Chamfer distance~($\downarrow$). The smallest and second smallest errors are labeled in bold and underlined. ``N/A'' denotes the experiment where a specific method cannot output reasonable shape estimation results.}
	\label{table:shape_est_syn}
	\centering
	\resizebox{.48\textwidth}{!}{
\begin{tabular}{lccccc}
	\toprule
	Method & {\sc Hedgehog } & {\sc Squirrel} & {\sc Snail} & {\sc David} & {\sc Dragon} \\
	\midrule
	 \nero & 5.39 & \underline{4.69} & 14.19 & 45.8  & 6.51  \\
	 \svolsdf & 7.33 & 5.33 & 16.8 & $\underline{5.12}$  &  N/A \\
	 \mvas & $\underline{5.37}$ & 5.72 & $\underline{8.01}$ & 7.01  & $\underline{6.48}$  \\
	 \pandora & 9.33 & 11.1 & 18.8 & 7.86  & 17.4 \\
	 \nersp~(Ours) & \textbf{3.43} & \textbf{4.55} &  \textbf{5.59} &  \textbf{4.16}  &  \textbf{4.77}  \\
	 \bottomrule
\end{tabular}	
}
\end{table}

\renewcommand{\imgw}{0.3}
\begin{figure}
	\small
	\resizebox{1\linewidth}{!}{
		\begin{tabular}{ccc}
			GT & \nero & \svolsdf \\  
			\includegraphics[width=\imgw\linewidth]{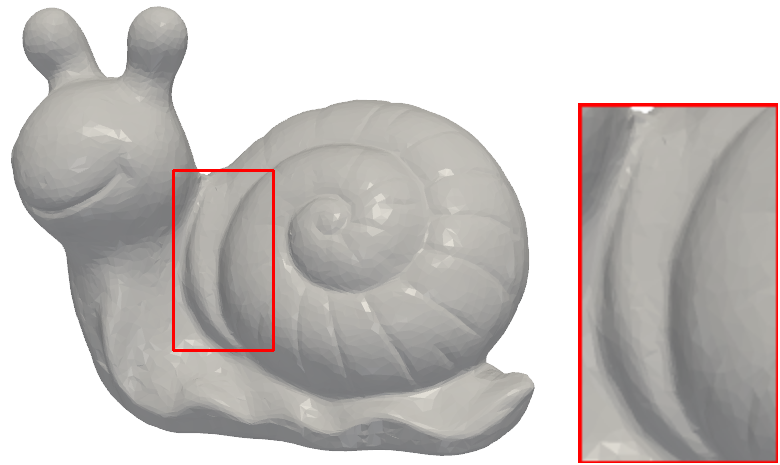}
			&\includegraphics[width=\imgw\linewidth]{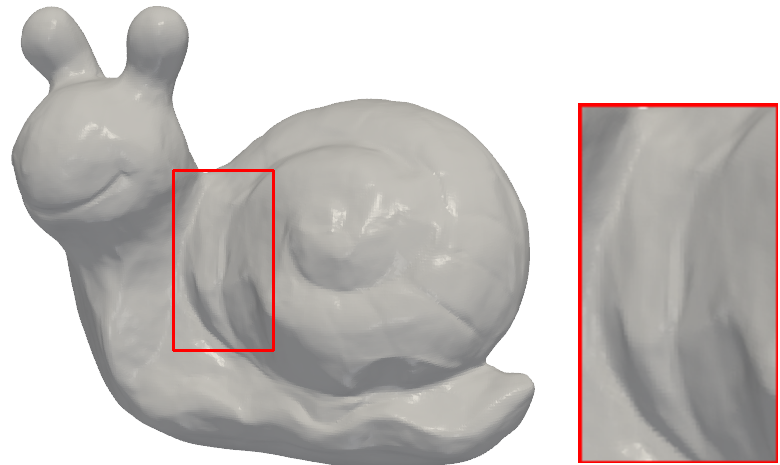}
			&\includegraphics[width=\imgw\linewidth]{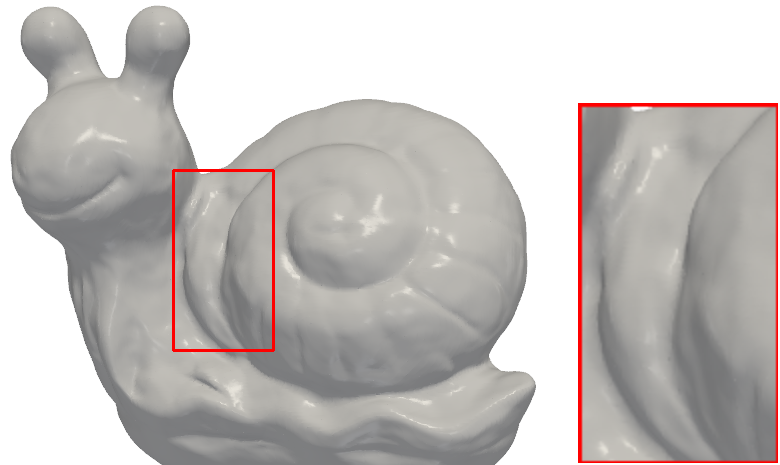}\\
			\mvas &	\pandora & \nersp~(ours)\\
			\includegraphics[width=\imgw\linewidth]{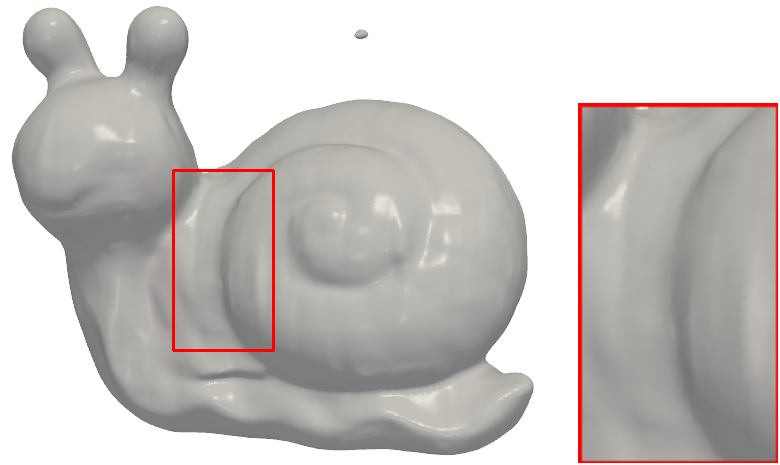}
			&\includegraphics[width=\imgw\linewidth]{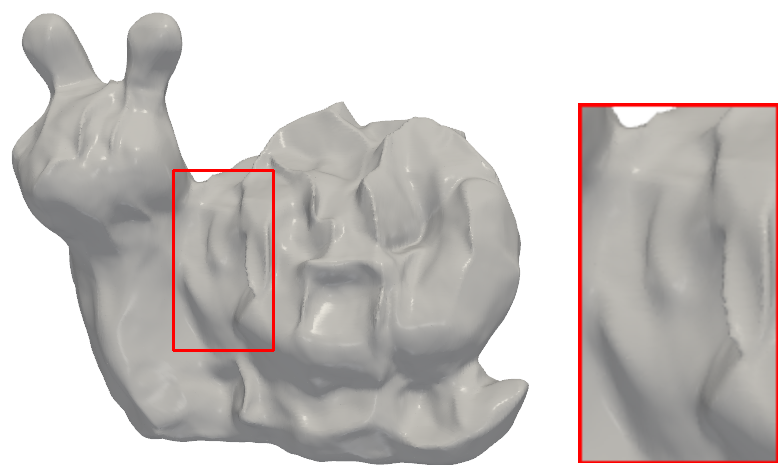}
			&\includegraphics[width=\imgw\linewidth]{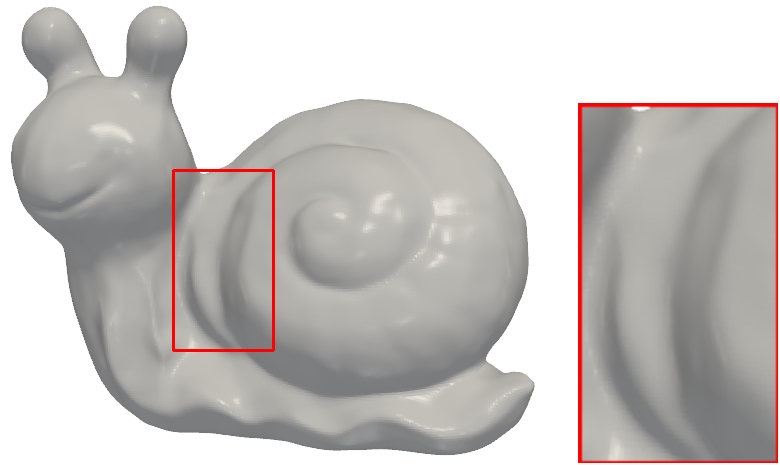}\\
			
		\end{tabular}
	}
	\caption{Qualitative evaluation on shape recoveries from $6$ sparse inputs. Our recovered shapes are closer to the GT, as highlighted in the closed-up views.
	}
	\label{fig:syn_shape_vis}
\end{figure}

Besides the evaluation of the reconstructed mesh, we also test the surface normal estimation results. 
As shown in \Tref{table:normal_est_syn}, we summarize the mean angular errors of estimated surface normals at $6$ views from different methods. 
Consistent with the evaluation results in \Tref{table:shape_est_syn}, \nersp achieves the smallest mean angular errors in average.
We also observed that the results from \nero, \mvas, and \pandora have larger errors on objects with fine details, such as {\sc David} and {\sc Dragon} objects. As an example, \mvas has the second smallest Chamfer distance shown in \Tref{table:shape_est_syn}, but the mean angular error is over $20^\circ$. One potential reason is existing methods output smooth shapes in the sparse views setting, where the surface details such as the flakes of the {\sc Dragon} are not well recovered. 
\subsection{Ablation study}
\begin{figure}
	\includegraphics[width=\linewidth]{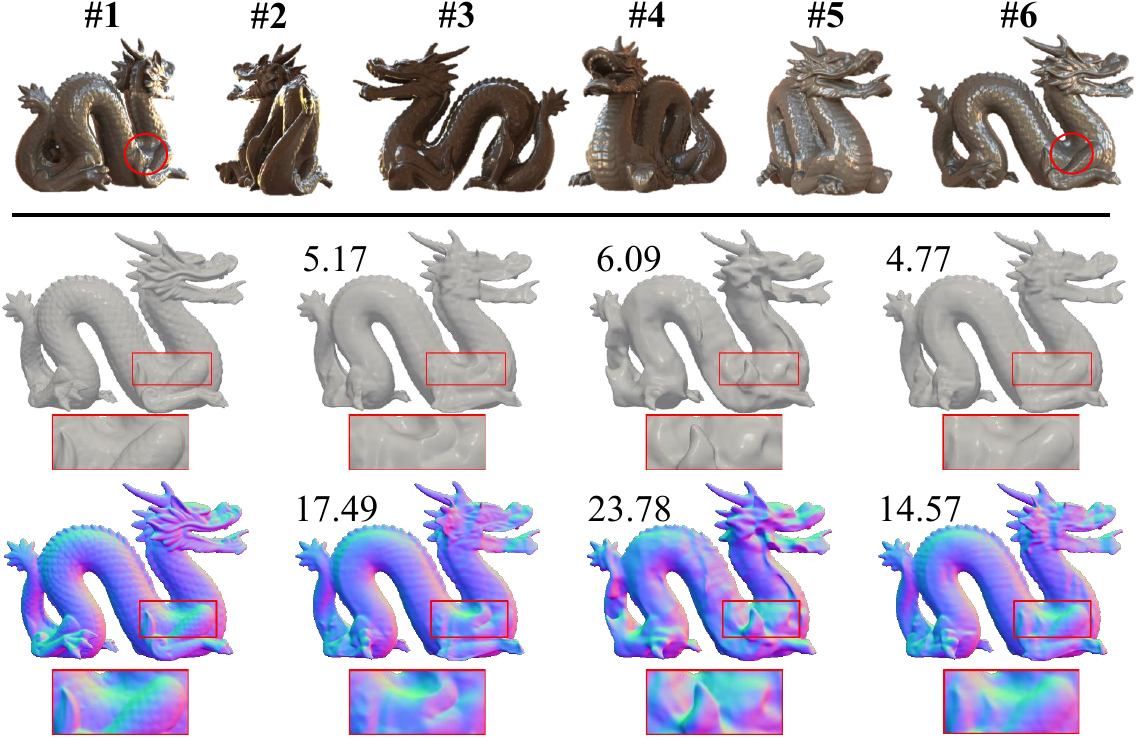}\\
	\makebox[0.1\textwidth]{GT}
	\makebox[0.1\textwidth]{\quad\quad  w/o $\mathcal{L}_p$}
	\makebox[0.1\textwidth]{\quad\quad\quad \quad w/o $\mathcal{L}_g$}
	\makebox[0.1\textwidth]{\quad\quad\quad\quad\quad\quad  Ours}
	
	\caption{Ablation study on different loss terms. The top and bottom rows visualize the estimated shape and surface normal, with the Chamfer distance and the mean angular error labeled on the top of each sub-figure, respectively.}
	\label{fig:ablation_loss}

\end{figure}

In this section, we conduct an ablation study to test the effectiveness of geometric and photometric cues.
Taking the {\sc Dragon} object as an example, we conduct our method with and without the photometric loss ${\mathcal{L}_p}$ and the geometric loss ${\mathcal L}_g$. As shown in \fref{fig:ablation_loss}, we plot the shape and surface normal estimations by disabling the different loss terms. 
Without the photometric loss, the shape ambiguity due to the sparse views occurs. As shown from the closed-up views, the shape near the leg part has a concave artifact, as there are only two visible views for this region, unable to formulate a unique solution for the shape merely based on the AoP maps~\cite{cao2023multi}. Without geometric loss, we also obtain distorted shape results as the sparse image observations are not sufficient to uniquely decompose the shape, reflectance, and illumination. By combining the photometric and geometric loss, our \nersp reduces the ambiguity of shape recovery and the estimated shape is closer to the GT, as highlighted in the closed-up views. 

\begin{table}
	\caption{Comparison on surface normal estimation on synthetic dataset evaluated by mean angular error~(MAE)~($\downarrow$).}
	\label{table:normal_est_syn}
	\resizebox{.49\textwidth}{!}{
		\begin{tabular}{lccccc}
			\toprule
			Method &  {\sc Hedgehog } & {\sc Squirrel} & {\sc Snail} & {\sc David} & {\sc Dragon} \\
			\midrule
			\nero & 9.14 & \underline{10.15} & 11.45 & 42.02  & \underline{24.22} \\
			\svolsdf & 11.26 & 13.28 & 7.59 & \underline{17.05}  & N/A  \\
			\mvas & \textbf{7.06} & 10.28 & \underline{6.19} & 21.86  & 24.29  \\
			\pandora & 19.75 & 23.52 & 16.54 & 21.88  & 28.82  \\
			\nersp~(Ours) & \underline{7.89} & \textbf{9.80} & \textbf{4.82} & \textbf{13.70}  & \textbf{18.03}  \\
			
			\bottomrule
		\end{tabular}	
	}
\end{table}

\subsection{Shape recovery on real data}
Besides the synthetic experiments shown in the previous section, we also evaluate our method on real-world datasets \pandoraD and \rmvp to test its applicability in real-world 3D reconstruction scenarios.

\begin{figure}
	\resizebox{\linewidth}{!}{
		\begin{tabular}{cccc}
			Polarized image & \nero & \svolsdf & \mvas
			\\
			\includegraphics[height=\imgw\linewidth]{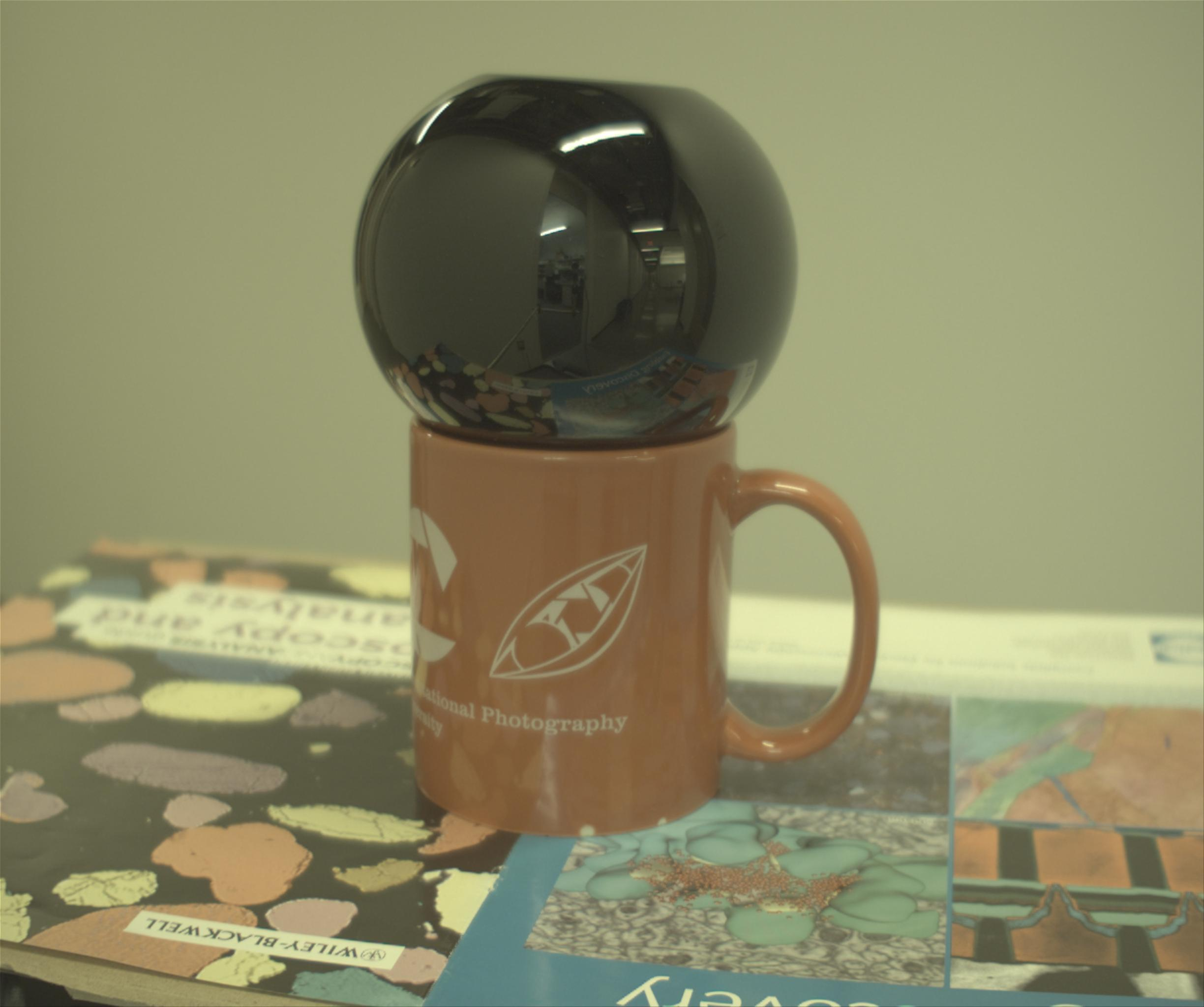}
			&\includegraphics[height=\imgw\linewidth]{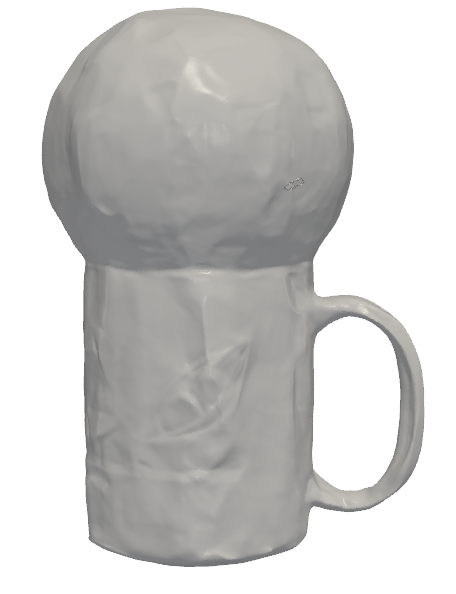}
			&\includegraphics[height=\imgw\linewidth]{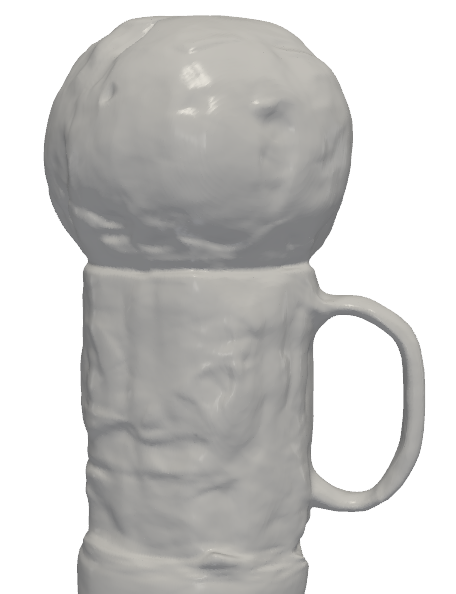}
			&\includegraphics[height=\imgw\linewidth]{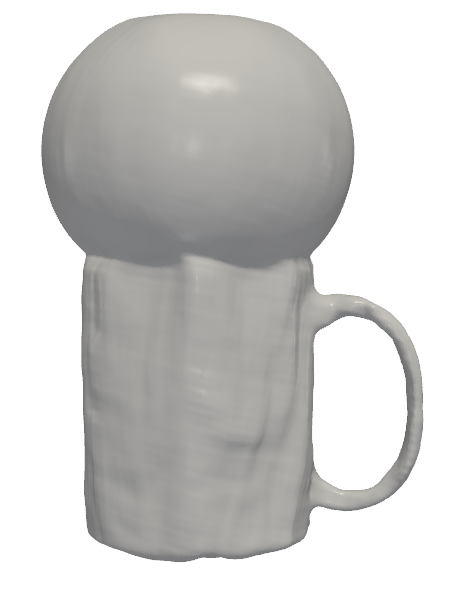}
			\\
			AoP map & \pandora & \nersp~(Ours) & 
			\\
			\includegraphics[height=\imgw\linewidth]{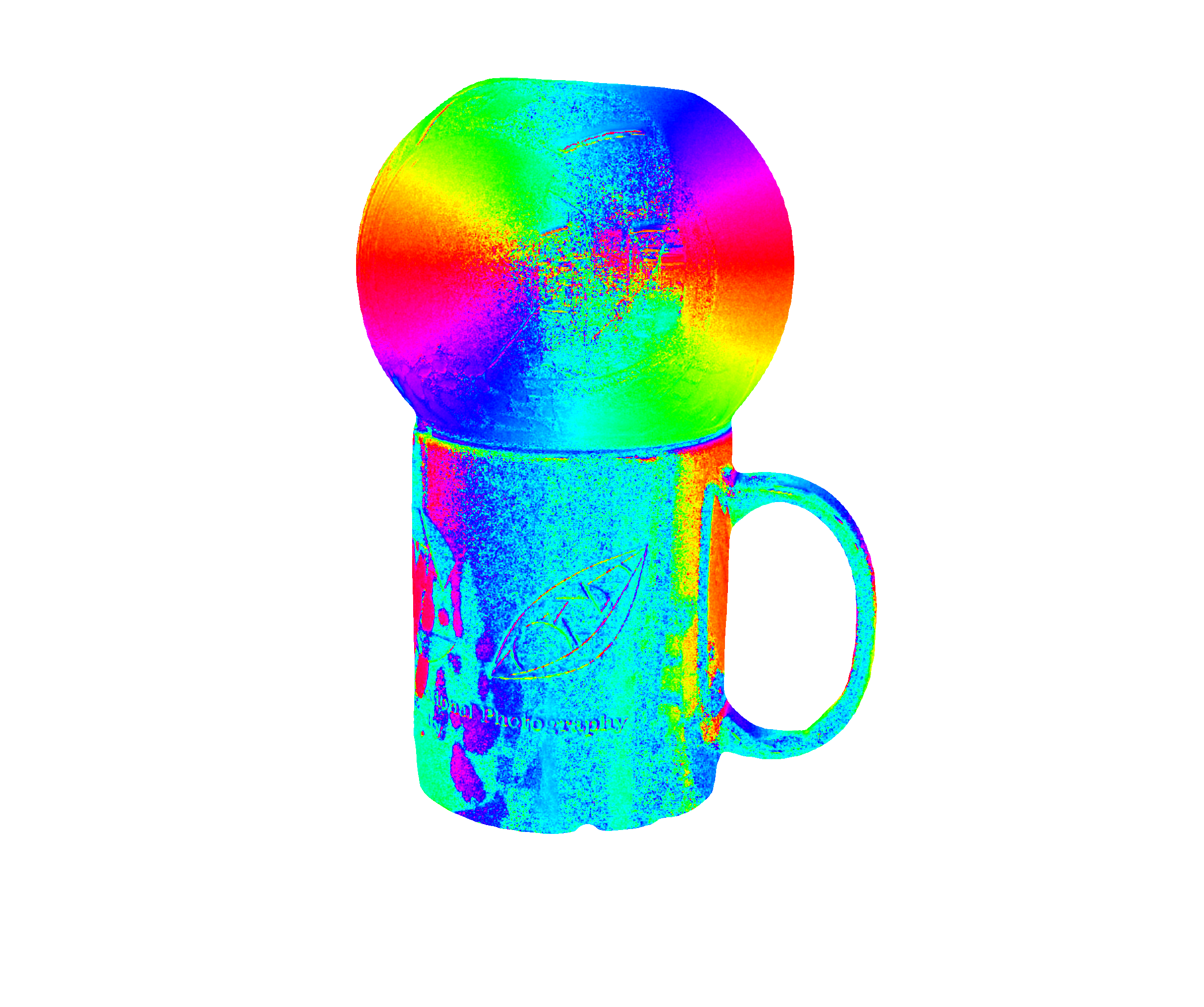}
			&\includegraphics[height=\imgw\linewidth]{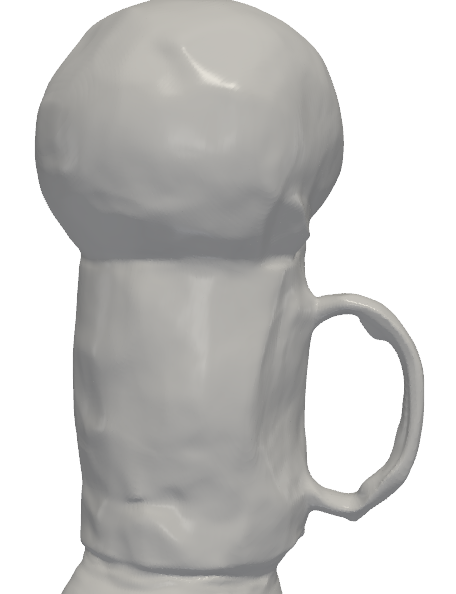}
			&\includegraphics[height=\imgw\linewidth]{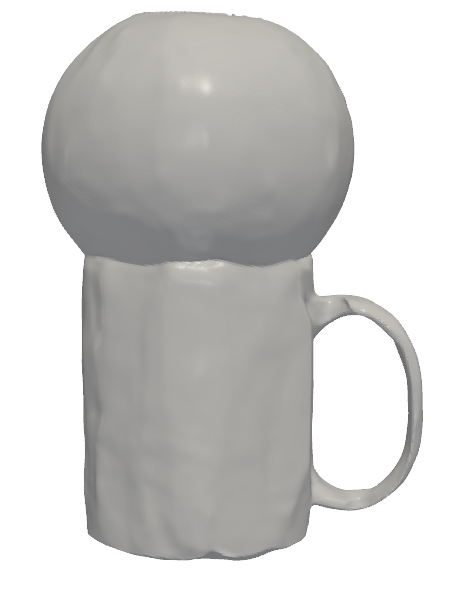}\\
			\midrule
			Polarized image & \nero & \svolsdf & \mvas
			\\
			\includegraphics[height=\imgw\linewidth]{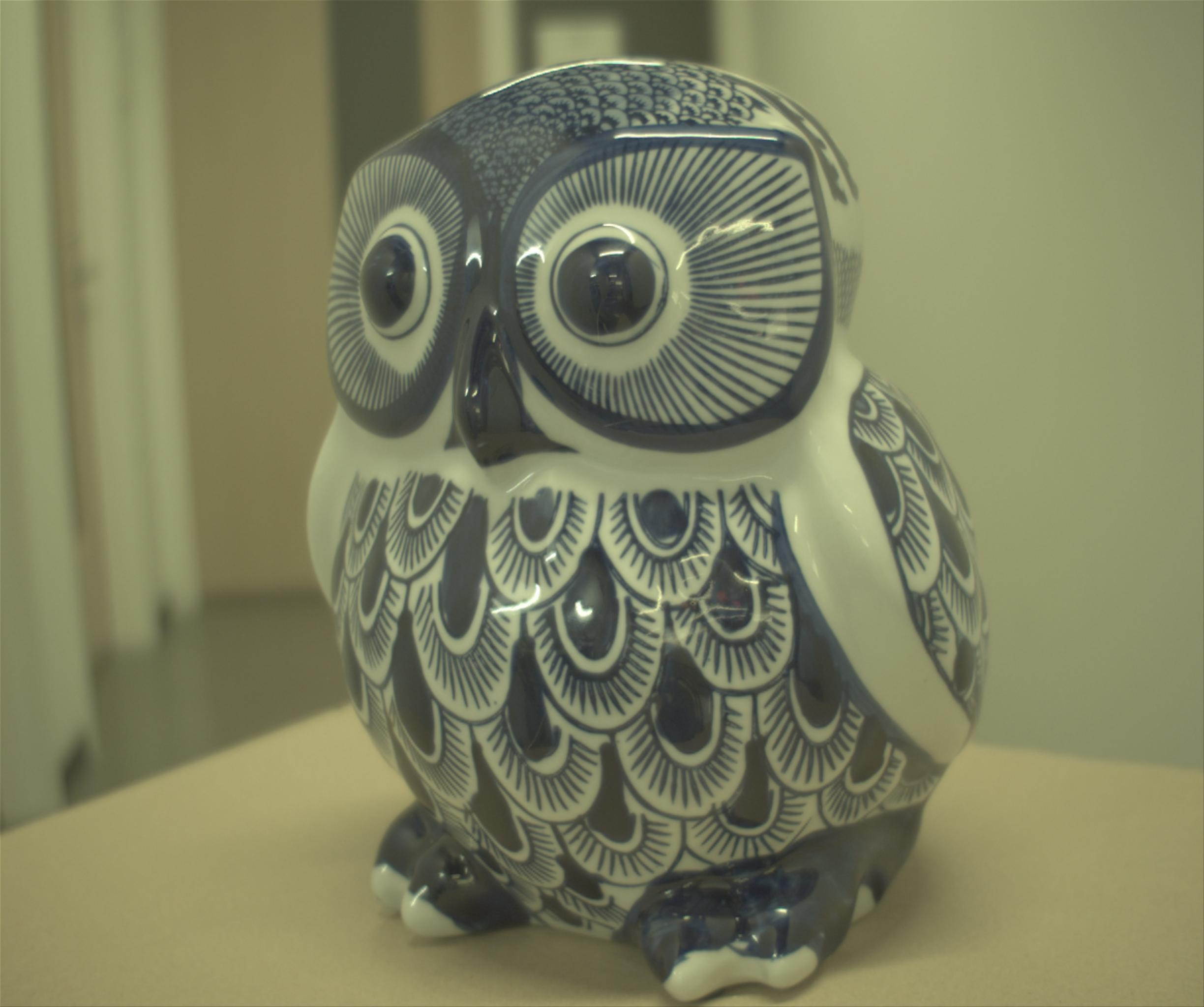}
			&\includegraphics[height=\imgw\linewidth]{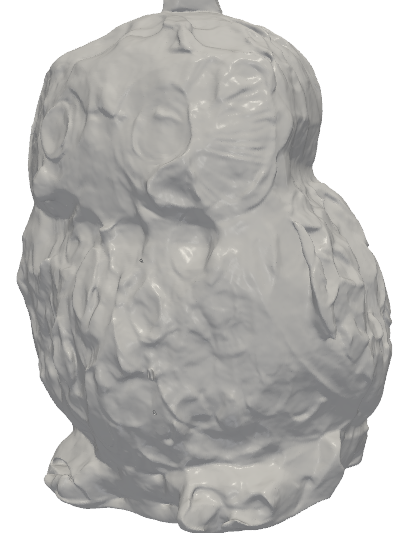}
			&\includegraphics[height=\imgw\linewidth]{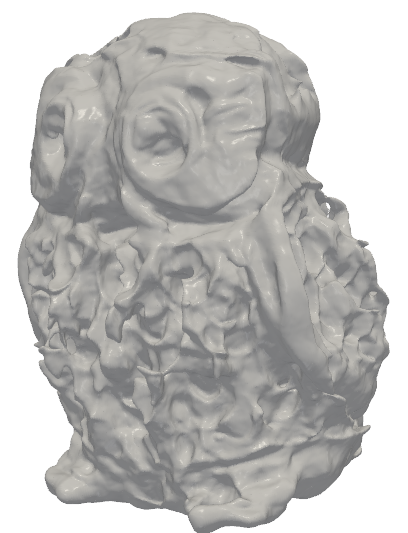}
			&\includegraphics[height=\imgw\linewidth]{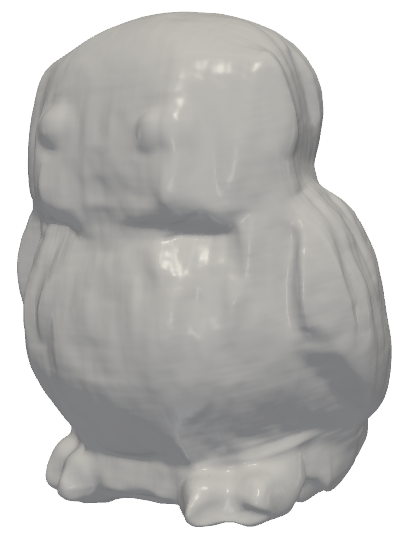}
			\\
			AoP map & \pandora & \nersp~(Ours) & 
			\\
			\includegraphics[height=\imgw\linewidth]{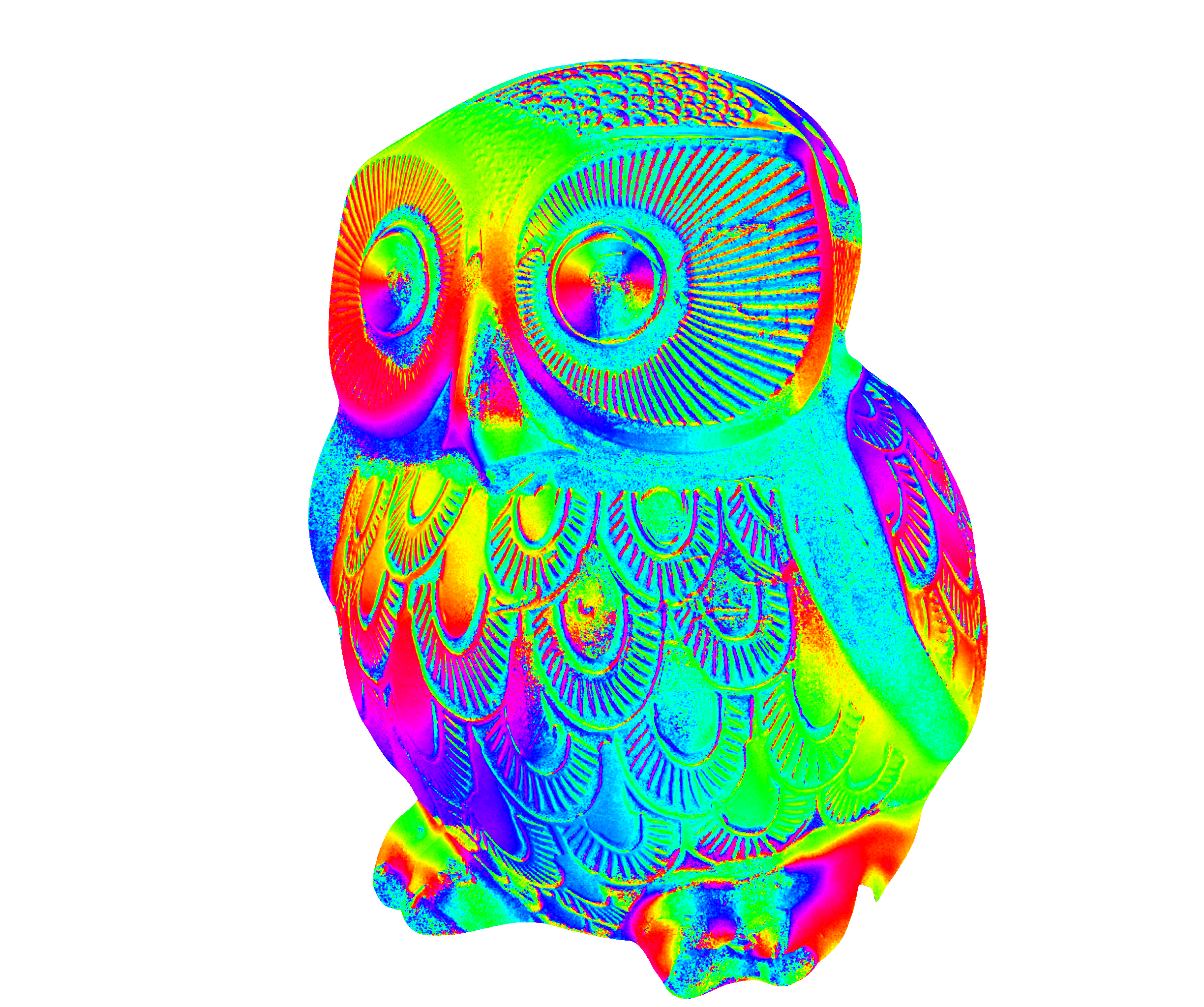}
			&\includegraphics[height=\imgw\linewidth]{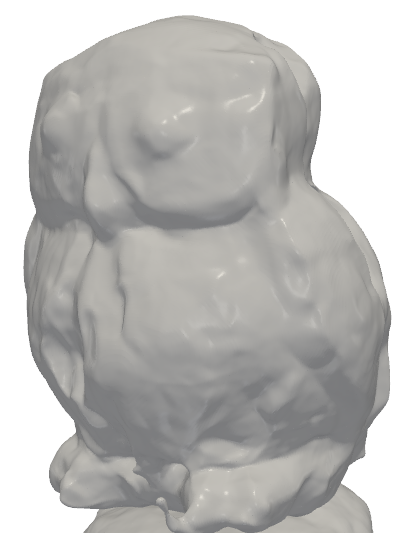}
			&\includegraphics[height=\imgw\linewidth]{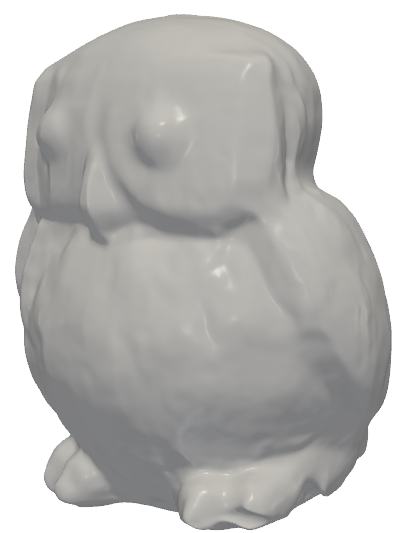}\\
		\end{tabular}
	}
	\caption{Qualitative evaluation on \pandoraD.}
	\label{fig:eval_real}

\end{figure}

\paragraph{Qualitative evaluation on \pandoraD.} As shown in \fref{fig:eval_real}, we provide qualitative evaluations on \pandoraD. Compared to the image appearance with the estimated results from \svolsdf and \nero, the shape is not fully disentangled from the reflectance, leading to bumpy surface shapes that are closely related to the reflectance texture. \mvas and \pandora have over-smoothed shape estimates or concave shape artifacts, due to addressing only geometric or photometric cues under the sparse capture setting. Our shape estimation results have no such shape artifacts and match the image observations closely.
\vspace{-1em}

\begin{table}
	\caption{Quantitative evaluation on \rmvp with Chamfer distance~($\downarrow$). Our method achieves the smallest error on average. }
	\label{table:eval_rmvp}
	\resizebox{\linewidth}{!}{
		\begin{tabular}{lccccc}
			\toprule
			Method & {\sc Dog} & {\sc Lion} & {\sc Frog} & {\sc Ball}  & Average \\ 
			\midrule
			\nero & \underline{9.11} & 10.74 & \textbf{6.21} & \underline{3.87}  & $\underline{7.48}$ \\
			\svolsdf & 9.93 & \underline{7.39} & 7.91 & 18.4 & 10.91 \\
			\mvas & 9.23 & 7.51 & 9.90 & 4.77  & 7.86\\
			\pandora & 14.3 & 15.04 & 11.27 & 3.96 & 11.14  \\
			\nersp~(Ours) & \textbf{8.80} & \textbf{5.18} & \underline{6.70} & \textbf{3.84}  & \textbf{6.13} \\
			\bottomrule
		\end{tabular}	
	}
	\vspace{-1em}
\end{table}

\begin{figure}
	\small
\resizebox{1\linewidth}{!}{
					\small
	\begin{tabular}{ccc}
		GT & \nero & \svolsdf \\  
		\includegraphics[width=\imgw\linewidth]{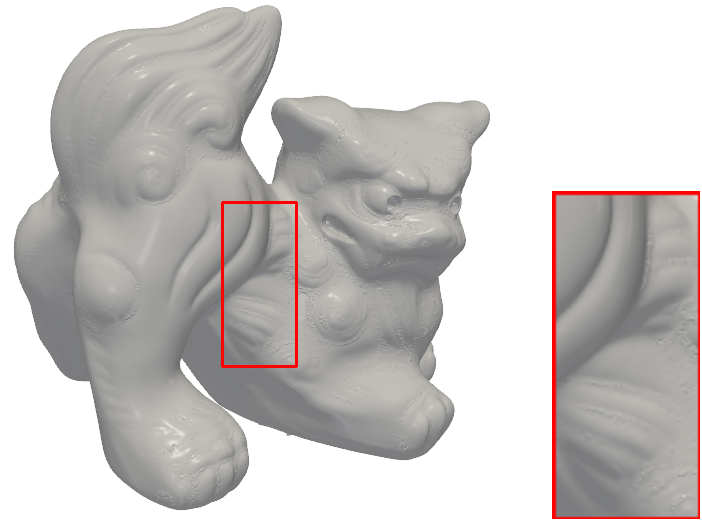}
		&\includegraphics[width=\imgw\linewidth]{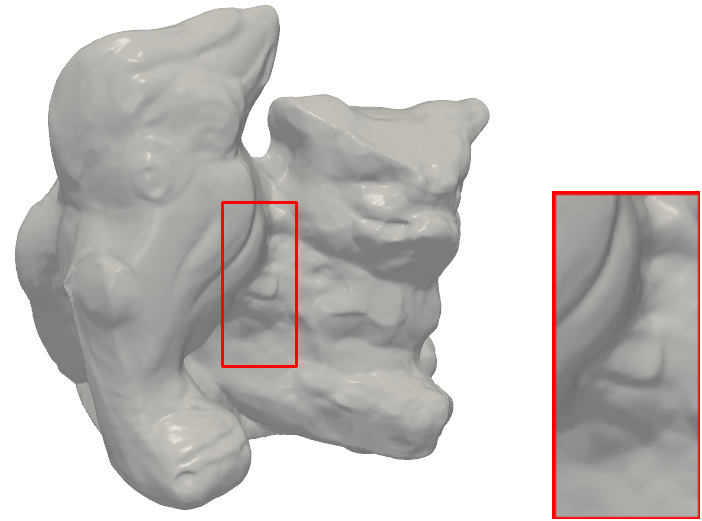}
		&\includegraphics[width=\imgw\linewidth]{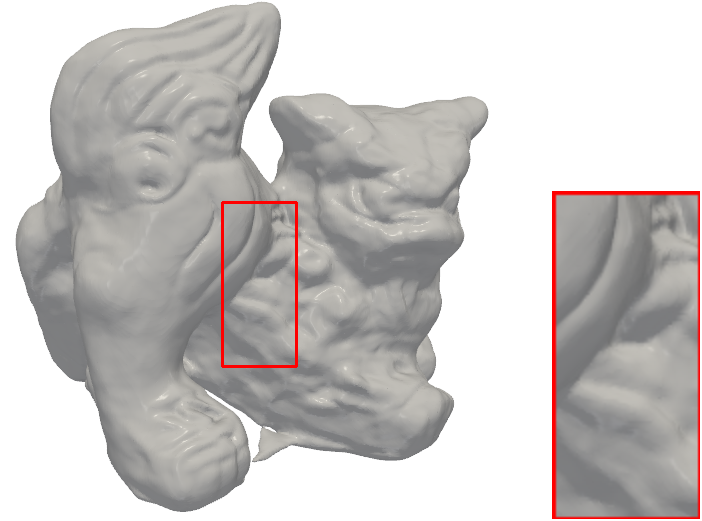}\\
		\mvas &	\pandora & \nersp~(ours)\\
		\includegraphics[width=\imgw\linewidth]{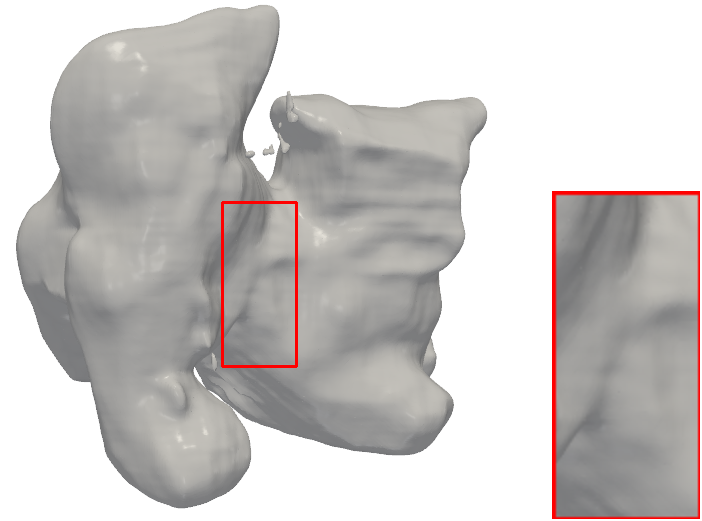}
		&\includegraphics[width=\imgw\linewidth]{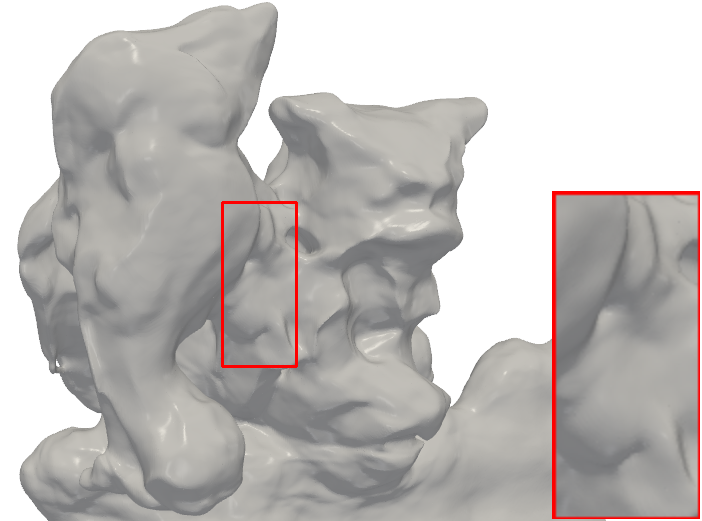}
		&\includegraphics[width=\imgw\linewidth]{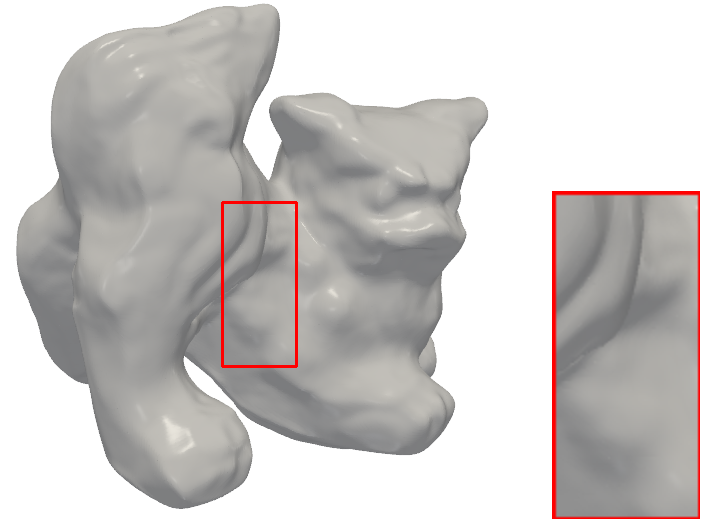}\\
		\midrule
		GT & \nero & \svolsdf \\ 
		\includegraphics[width=\imgw\linewidth]{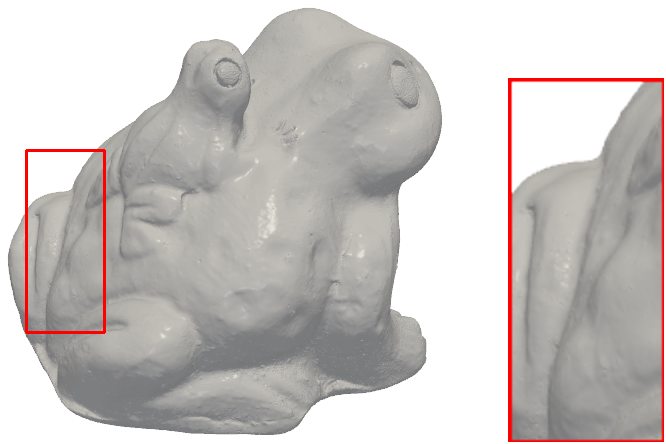}
		&\includegraphics[width=\imgw\linewidth]{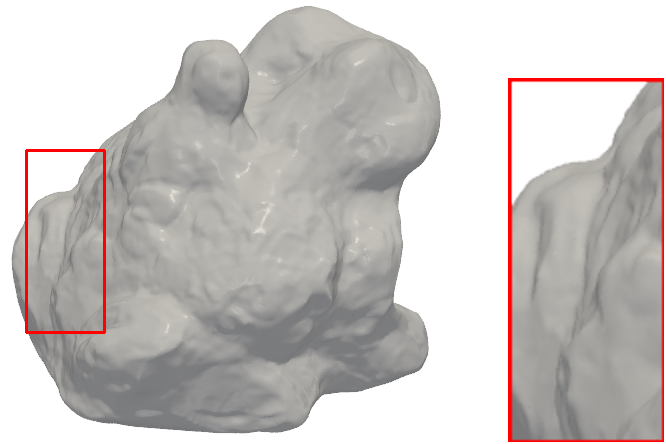}
		&\includegraphics[width=\imgw\linewidth]{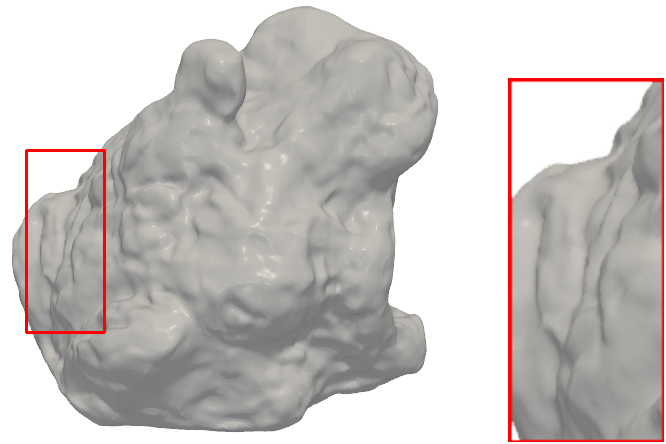}\\
		\mvas &	\pandora & \nersp~(ours)\\
		\includegraphics[width=\imgw\linewidth]{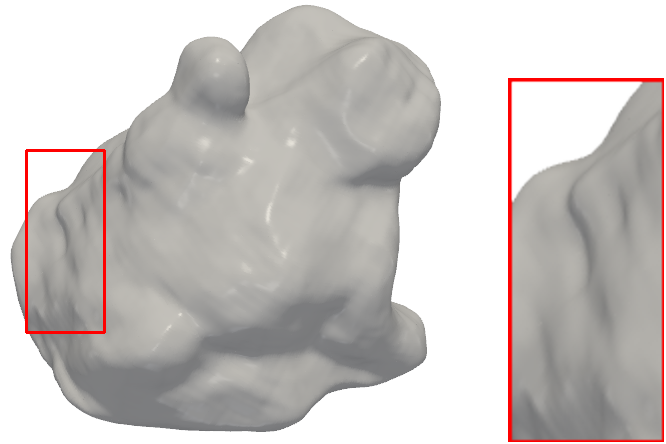}
		&\includegraphics[width=\imgw\linewidth]{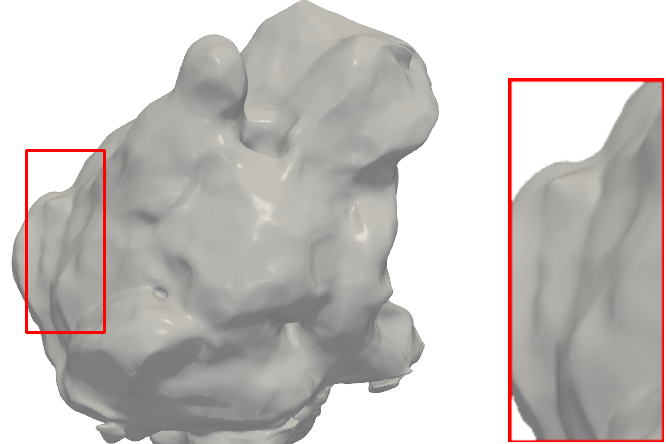}
		&\includegraphics[width=\imgw\linewidth]{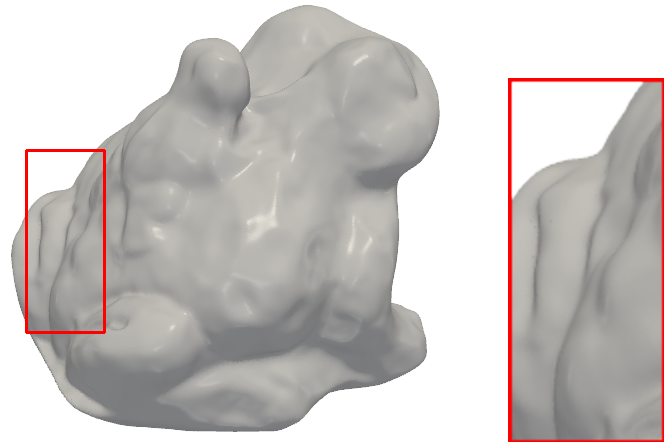}\\
	\end{tabular}
}
	\caption{Comparison on shape recoveries on \rmvp dataset.}
	\label{fig:eval_real_shape_rmvp}
	\vspace{-1em}
\end{figure}

\paragraph{Quantitative evaluation on \rmvp.}
As shown in \Tref{table:eval_rmvp}, we present a quantitative evaluation on \rmvp based on Chamfer distance. Consistent with the synthetic experiment, our \nersp achieves the smallest estimation error in average. The visualized shapes shown in \fref{fig:eval_real_shape_rmvp} further reveal that reflective surfaces are challenging to \svolsdf for disentangling the shape from reflectance, as highlighted by the bumpy surface of the {\sc Frog} object in the closed-up views. \nero and \pandora have similar estimation error with us on the simple {\sc Ball} object. For complex shapes like {\sc Lion}, distorted shape recoveries are obtained from these methods due to the sparse views setting, while ours are closer to the GT meshes, demonstrating the effectiveness of our method on real-world reflective surface reconstruction under sparse inputs.

\section{Conclusion}
We propose \nersp, a neural 3D reconstruction method for reflective surfaces under sparse polarized images. 
Due to the challenges of shape-radiance ambiguity and complex reflectance, existing methods struggle with either reflective surfaces or sparse views and cannot address both problems with RGB images. We propose to use polarized images as input. By combining the geometric and photometric cues extracted from polarized images, we reduce the solution space of the estimated shape, allowing for the effective recovery of reflective surface with as few as $6$ views, as demonstrated by publicly available and our own datasets.
\vspace{-0.8em}
\paragraph{Limitation}
The inter-reflections and polarized environment light are not considered in this work, which could influence the shape reconstruction accuracy. 
We noticed a most recent work NeISF~\cite{li2023neisf} focusing on this topic, and we are interested in combining our sparse shot merit with this work in the future.

\vspace{-0.8em}
\paragraph{Acknowledgment}
This work was supported by the Beijing Natural Science Foundation Project No. Z200002, the National Nature Science Foundation of China (Grant No.~62136001, 62088102, 62225601, U23B2052), the Youth Innovative Research Team of BUPT No. 2023QNTD02, and the JSPS KAKENHI (Grant No. JP22K17910 and JP23H05491). We thank Youwei Lyu for insightful discussions.

{
    \small
    \bibliographystyle{ieeenat_fullname}
    \bibliography{heng}
}

\end{document}